%% file: main.tex
\documentclass[sigplan]{acmart}

\usepackage{xcolor}
\usepackage{subcaption} 
\usepackage{graphicx}
\usepackage{amsmath}
\usepackage{enumerate}
\usepackage{calrsfs}
\usepackage{booktabs}
\usepackage{multirow}
\usepackage[noend]{algpseudocode}
\usepackage{algorithm}
\usepackage{mathtools}
\usepackage{xfrac}
\usepackage{pifont}
\usepackage{makecell}
\usepackage{pdflscape}
\usepackage{geometry}
\usepackage{multirow}
\usepackage{caption}
\usepackage{lipsum}
\usepackage{wrapfig}
\usepackage{enumitem} 

\AtBeginDocument{%
  }
    
\begin{document}

\acmYear{2024}\copyrightyear{2024}
\setcopyright{rightsretained}
\acmConference[ACM FAccT '24]{ACM Conference on Fairness, Accountability, and Transparency}{June 3--6, 2024}{Rio de Janeiro, Brazil}
\acmBooktitle{ACM Conference on Fairness, Accountability, and Transparency (ACM FAccT '24), June 3--6, 2024, Rio de Janeiro, Brazil}
\acmDOI{10.1145/3630106.3658954}
\acmISBN{979-8-4007-0450-5/24/06}

\title{Mitigating Group Bias in Federated Learning for Heterogeneous Devices}

\author{Khotso Selialia}
\affiliation{%
  \institution{University of Massachusetts Amherst}
  \city{Amherst}
  \state{Massachusetts}
  \country{USA}}
\email{}

\author{Yasra Chandio}
\affiliation{%
  \institution{University of Massachusetts Amherst}
  \city{Amherst}
  \state{Massachusetts}
  \country{USA}}
\email{}

\author{Fatima M. Anwar}
\affiliation{%
  \institution{University of Massachusetts Amherst}
  \city{Amherst}
  \state{Massachusetts}
  \country{USA}}
\email{}

\renewcommand{\shortauthors}{Selialia et al.}

\begin{abstract}
  Federated learning is emerging as a privacy-preserving model training approach in distributed edge applications. As such, most edge deployments are heterogeneous in nature, i.e., their sensing capabilities and environments vary across deployments. This edge heterogeneity violates the independence and identical distribution (IID) property of local data across clients. It produces biased global models, i.e., models that contribute to unfair decision-making and discrimination against a particular community or a group. Existing bias mitigation techniques only focus on bias generated from label heterogeneity in non-IID data without accounting for domain variations due to feature heterogeneity.

Our work proposes a group-fair FL framework that minimizes group-bias while preserving privacy. Our main idea is to leverage average conditional probabilities to compute a cross-domain group \textit{importance weights} derived from heterogeneous training data to optimize the performance of the worst-performing group using a modified multiplicative weights update method. Additionally, we propose regularization techniques to minimize the difference between the worst and best-performing groups while ensuring through our thresholding mechanism to strike a balance between bias reduction and group performance degradation. Our evaluation of image classification benchmarks assesses the fair decision-making of our framework in real-world settings.
\end{abstract}




\maketitle

\section{Introduction}
\label{sec:introduction}
\input{1-introduction}
\section{Background and Related Work}
\label{sec:background}
\input{2-background}

\section{Preliminary Study}
\label{sec:preliminary-study}
\input{3-prelim_study}
\section{Methodology} 
\label{sec:methodology}
\input{4-methodology}
\section{Evaluation}
\label{sec:evaluation}
\input{5-evaluation}
\section{Conclusion and Future Work}
\label{sec:conclusion}
\input{6-conclusion}

\begin{acks}
This material is based upon work supported by the National Science Foundation under grant number $2237485$.
\end{acks}

\bibliographystyle{ACM-Reference-Format}
\bibliography{sample-base}

\end{document}

%% file: 1-introduction.tex
\begin{figure*}[t]
    \centering
    \includegraphics[width=\textwidth]{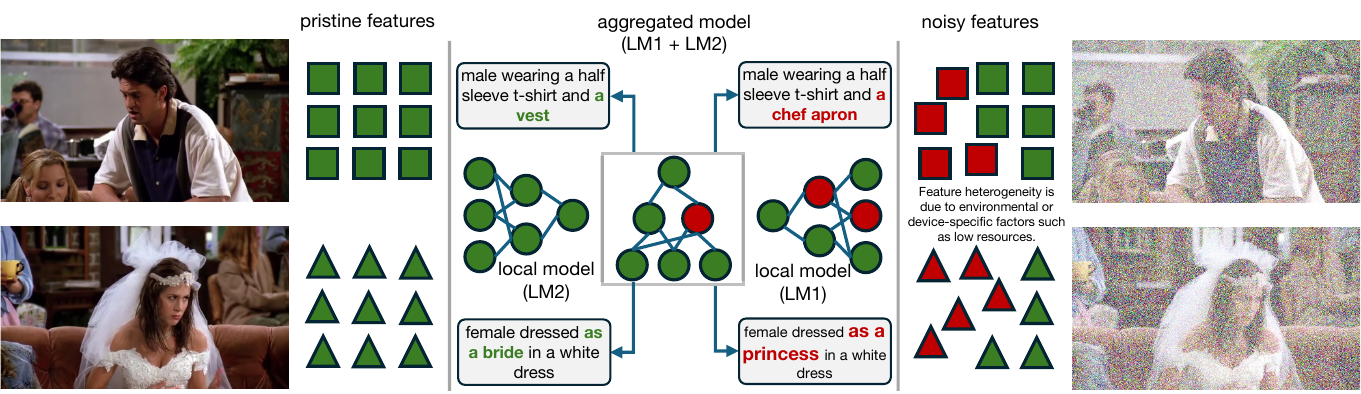}
    \caption{Illustrating the adverse effects of \emph{feature heterogeneity (noise)} and its bias impact on image classification data \cite{lee2019context} on an example language model (LM) in FL settings. The global LM, engaging in image captioning based on features from multiple clients, shows higher performance for images without distortions compared to those with a shift in feature distributions. This emphasizes the intricate interplay of feature heterogeneity and bias in FL, highlighting the influence of heterogeneous client datasets on the model's outcome.}
    \label{fig:problem}
\end{figure*}

Federated learning (FL) is a privacy-preserving machine learning (ML) technique wherein local models are trained on decentralized edge devices (\emph{clients}) and subsequently aggregated at the server to form a \emph{global model}. This approach alleviates the need for raw data transfers and ensures data privacy, making it particularly well-suited for applications with privacy sensitivities, such as medical diagnosis \cite{qayyum2022collaborative, feki2021federated, ku2022privacy}, next-character prediction\cite{sun2022fedsea}, activity recognition~\cite{sozinov2018human, ouyang2021clusterfl, ek2020evaluation}, and human emotion recognition~\cite{4287274, article1, 5740836}, where preserving data security is imperative. Despite its merits, there is a growing concern regarding FL models, as they exhibit exceptional performance for certain groups while simultaneously underperforming for others (e.g., providing accurate image captioning for pristine group images than noisy group images as shown in Figure \ref{fig:problem}). A group categorizes data based on attributes such as race, gender, class, or label~\cite{chen2022fair}.

Group biases and discriminatory practices threaten societal well-being, undermining public confidence in ML models and their applications~\cite{chen2022fair}. Research shows racial bias in electronic health records, especially in medical analysis, potentially causing treatment disparities for minority groups~\cite{sun2022negative}. Biased models often result from label heterogeneity in non-IID data across clients, as discussed in works like \cite{papadaki2022minimax, mohri2019agnostic}, arising from diverse label distributions tied to data collection device environments. For example, certain geo-regions may have varying label distributions, reducing training data volume for specific groups~\cite{chen2018my, hsieh2020non}.

Our work highlights \emph{feature noise heterogeneity} as a significant source of group bias in FL models, stemming from varied noise-influenced features due to domain differences, especially in heterogeneous devices~\cite{lyu2020collaborative}. Heterogeneity leads to distinct feature distributions in local client data. For example, low-quality sensors on some devices introduce distortion like Gaussian noise, resulting in different feature distributions compared to high-quality sensor devices~\cite{liu2020collabar}. This inherent feature noise causes shifts in group data moments, which are statistical properties such as mean and variance within a group in a dataset~\cite{khani2020feature}, influencing biased model outcomes.

Previous FL research introduces Disparate Learning Processes (DLPs) to tackle bias and fairness issues. Examples of DLPs include \emph{in-processing} methods like \cite{mohri2019agnostic, deng2020distributionally, ro2021communication, li2019fair, li2020tilted, hu2020fedmgda+, chu2021fedfair, cui2021addressing, du2021robust, du2021fairness, ezzeldin2023fairfed, zeng2021improving, galvez2021enforcing, yue2021gifair, papadaki2022minimax, zhang2021unified} and \emph{Robustness and generalization} strategies such as \cite{lee2022preservation, karimireddy2020scaffold}. In-processing techniques modify learning to include group fairness constraints, while robustness and generalization enhance model resilience in diverse data settings. However, DLPs don't ensure fairness in settings with feature heterogeneity, especially due to feature noise, as they don't address misaligned moments in feature distributions \cite{khani2020feature}. For DLPs that use "reweighting" with \emph{importance weights} to adjust the model's objective function, their effectiveness relies on suitable importance weight selection \cite{byrd2019effect}. Importance weights prioritize specific groups or features during training to mitigate biases and enhance fairness \cite{byrd2019effect}. If not chosen carefully or not aligned with genuine sources of bias, these weights can lead to continued unfairness \cite{byrd2019effect}. We propose using weights derived from noisy feature data for more efficient debiasing in FL models affected by feature noise. This work introduces learnable importance weights from heterogeneous data features to enhance fairness in training, utilizing the \emph{multiplicative weight update (MW)} method \cite{arora2012multiplicative} for better fairness based on feature characteristics, especially considering data characteristics with feature noise. Our approach is inspired by insights from social science, particularly addressing discrimination as a health disparity determinant \cite{krieger2012methods}. By incorporating learnable importance weights, we aim to mitigate biases across demographic groups, contributing to a more equitable FL framework.

The efficacy of \emph{importance weighting} diminishes due to exploding weight norms from the empirical risk scaling with importance weights, especially in large models, risking overfitting~\cite{byrd2019effect}. To tackle this, we propose using neural network regularization techniques~\cite{nusrat2018comparison} in \emph{Multiplicative Weight update with Regularization (MWR)} to mitigate \emph{group bias}. Additionally, methods using \emph{importance weighting} may introduce unfairness by overly emphasizing poorly-performing groups, potentially reducing the performance of better-performing groups to minimize overall variability \cite{diana2021minimax}. To address this issue, we present a heuristic approach for deriving \emph{importance weights} that mitigate group bias while maintaining a performance threshold for better-performing groups, preventing their performance from dropping below a desirable level. We summarize our contributions below:
\begin{itemize}[noitemsep, leftmargin=*]
    \item \textbf{Enabling Privacy-preserving Group Fairness}: We highlight the notion of group fairness across clients in FL settings and propose a \emph{Multiplicative Weight (MW)} update method to mitigate bias due to feature heterogeneity. Our approach requires an estimate of the global group importance weights, which we compute as a mixture of cross-domain likelihood estimates of heterogeneous local data across clients in a privacy-preserving manner.   
    \item \textbf{Ensuring Optimality through Regularization}: We extend our approach by incorporating the L1 regularization technique to increase its effectiveness in mitigating group bias, which we call \emph{MWR}. It combats diverging weight norms that fail to converge to a model that optimizes worst group performance.
    \item \textbf{Satisfying Worst- and Best-group Performance}: We ensure that \emph{MWR} optimizes the performance of the worst-performing group while also keeping the performance of the best-performing group above a desirable threshold.
    \item \textbf{Implementation and Evaluation}: We implement and evaluate the \emph{MWR} method against existing bias-mitigation techniques on commonly used state-of-the-art image classification FL benchmark datasets (CIFAR10~\cite{krizhevsky2009learning}, MNIST~\cite{lecun1998gradient},  FashionMNIST~\cite{xiao2017fashion}, USPS~\cite{hull1994database}, SynthDigits~\cite{ganin2015unsupervised}, and MNIST-M ~\cite{ganin2015unsupervised}). Our findings show that \emph{MWR} outperforms baseline methods, boosting the accuracy of the worst group's performance up to 41\% without substantially degrading the best group's performance.
\end{itemize}

%% file: 2-background.tex
\subsection{Bias in Machine Learning.} 
Bias in ML refers to a model favoring specific individuals or groups, leading to unfair outcomes \cite{mehrabi2021survey}. Common sources of bias in centralized learning include prejudice, underestimation, and negative legacy \cite{abay2020mitigating, chen2018my, madras2019fairness}. 
Techniques such as pre-processing, in-processing, and post-processing \cite{grgic2018beyond, feldman2015certifying, kamishima2012fairness} have effectively mitigated centralized learning bias. However, applying centralized learning techniques in FL is challenging due to privacy concerns, requiring access to features across clients and risking data privacy compromise.

\subsection{Bias Metrics}
In FL, \emph{group bias} is assessed through three dimensions: 1) aiming for equal opportunities by evaluating the performance discrepancy in True Positive Rates (\emph{TPR}) between groups \cite{wan2021modeling, poulain2023improving}; 2) optimizing the Worst-case TPR (\emph{WTPR}) for each group \cite{martinez2020minimax, poulain2023improving}; 3) minimizing the standard deviation of TPR (\emph{TPSD}) to ensure fairness across groups \cite{yue2021gifair, poulain2023improving}. The choice of TPR as a performance metric of in assessing group group fairness aligns our approach with recent advancements in bias mitigation literature \cite{poulain2023improving}. This decision stems from recognizing the critical importance of fairly detecting true positives, which cannot be addressed solely by relying on accuracy. While our primary focus is on achieving fairness with a \emph{minimax} property (optimizing \emph{WTPR} outcome within each group), we evaluate using various fairness metrics to ensure versatility and broad support.
\subsection{Bias Mitigation} 
The bias mitigation work falls mainly into four categories, including:
$1)$ \emph{Client-fairness} techniques \cite{mohri2019agnostic, deng2020distributionally, ro2021communication, li2019fair, li2020tilted, hu2020fedmgda+},
$2)$ \emph{Group-fairness} techniques \cite{chu2021fedfair, cui2021addressing, du2021robust, du2021fairness, ezzeldin2023fairfed, zeng2021improving, galvez2021enforcing, yue2021gifair, papadaki2022minimax, zhang2021unified},
$3)$ \emph{Collaborative Fairness} techniques \cite{fan2022improving, lyu2020collaborative, nagalapatti2021game, zhang2020fairfl}, and
$4)$ \emph{Robustness and Generalization} techniques \cite{reisizadeh2020robust, karimireddy2020scaffold, lee2022preservation}.
\begin{figure*}
     \centering
      \begin{subfigure}[b]{0.45\linewidth}
         \centering
         \includegraphics[width=\textwidth]{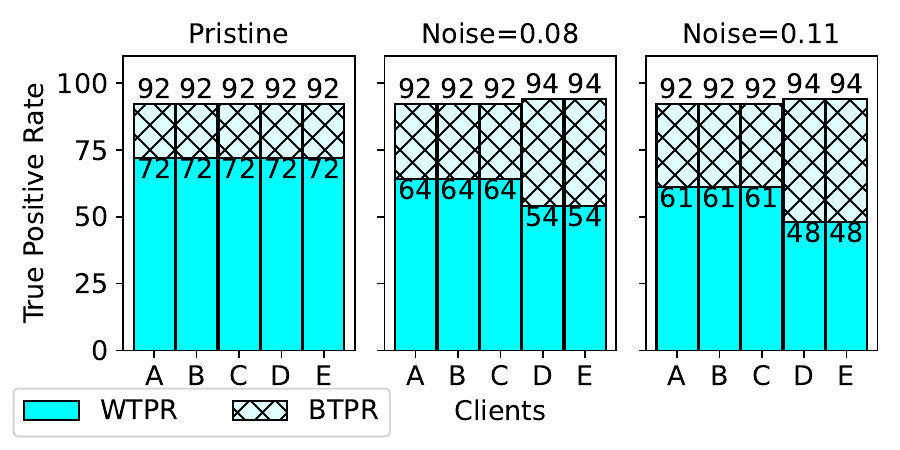}
         \vspace{-0.3cm}
         \caption{CIFAR10 dataset}
         \label{fig:cifarweights}
     \end{subfigure}
     \begin{subfigure}[b]{0.45\linewidth}
         \centering
         \includegraphics[width=\textwidth]{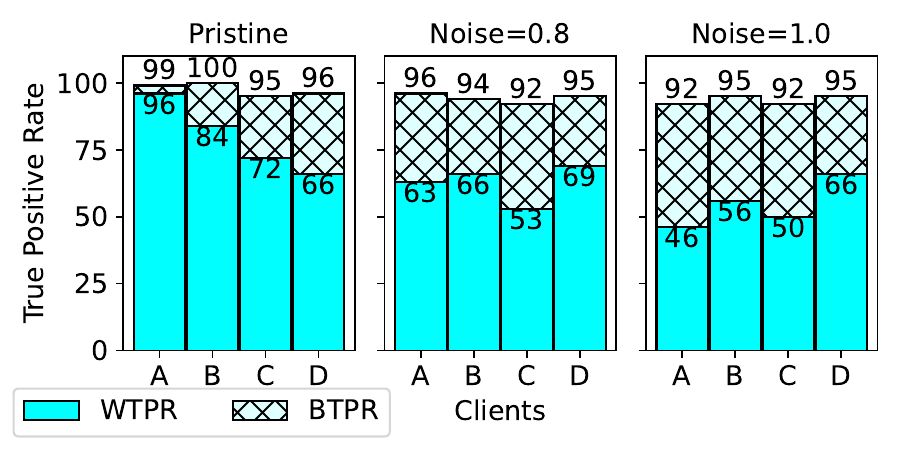}
         \vspace{-0.3cm}
         \caption{DIGITS dataset}
         \label{fig:digitsweights}
     \end{subfigure}
         \vspace{-0.1cm}
    \caption{Varied noise levels in CIFAR10 and DIGITS datasets. The notation "$\text{Noise}=x$" denotes the introduction of Gaussian noise with variance $x$," specifically applied to clients $D$ and $E$ in CIFAR10 and clients $A$ and $B$ in DIGITS.}
        \label{fig:prelim-bias}
\end{figure*}

\noindent\textbf{\emph{Client fairness}} targets the development of algorithms leading to models that exhibit similar performance across different clients \cite{li2019fair}. On the other hand, \textbf{\emph{group fairness}} requires the model to perform similarly on different demographic groups \cite{yue2021gifair}. Many state-of-the-art fairness techniques in FL, focusing on \emph{client fairness} and \emph{group fairness}, use \emph{in-processing} methods to modify the learning process or objective function by incorporating fairness constraints~\cite{yue2021gifair}. In-processing involves assigning weights to the objective function from different clients or groups during training to balance the influence of the model on different groups or clients. For instance, AFL \cite{mohri2019agnostic} optimizes the combination of worst-weighted losses from local clients, proving resilient to data with an unknown distribution. q-FFL \cite{li2019fair} reweights loss functions to give higher weights to devices with poorer performance, addressing challenges in fair resource allocation in computer networks. TERM  handles outliers and class imbalance by tilting the loss function with a designated tilting factor~\cite{li2020tilted}. GIFAIR-FL~\cite{yue2021gifair} introduces a regularization term, regarded as loss function weighting, to guide the optimizer towards group fair solutions. Despite the benefits, in-processing techniques face challenges, particularly sensitivity to outliers and dependence on the choice of reweighting schemes. If importance weights do not align well with data characteristics, outliers introduced by noise can have a significant impact, leading to biases. Feature noise may cause alterations in the distribution of features among groups, inducing discrepancies and bias in statistical properties.
 
\noindent\textbf{\emph{Collaborative Fairness} }methodologies propose compensating each client's performance based on their contribution to learning the global model, intending to align rewards with individual client input. This approach entails providing more rewards to highly contributing clients, thereby encouraging active participation in FL. Conversely, offering lower rewards helps prevent free-riders, ensuring a fair distribution of incentives \cite{lyu2020collaborative}. It is important to note that while we discuss \emph{Collaborative Fairness}, here does not specifically address mitigating group bias in FL, as these techniques do not inherently focus on improving group performances.

\textbf{\emph{Robustness and Generalization}} techniques address distributional shifts in user data. For instance, FedRobust \cite{reisizadeh2020robust} trains a model to handle worst-case affine shifts, assuming that each client can express its data distribution as an affine transformation of a global distribution, focusing on group fairness. However, FedRobust requires sufficient data for each client to estimate the local worst-case shift, impacting global model performance when this condition is unmet. FedNTD tackles \emph{catastrophic forgetting} distillation \cite{hinton2015distilling} but may not fully handle bias from feature noise. SCAFFOLD \cite{karimireddy2020scaffold} addresses client drift in heterogeneous data by estimating update directions. However, SCAFFOLD may not correct moments in noisy feature distributions. In contrast, we use importance weights from noisy features to prioritize disadvantaged groups during training, enhancing fairness by indirectly correcting misaligned moments.

%% file: 3-prelim_study.tex
This section analyzes \emph{group-bias} arising from heterogeneous feature distributions within local data across clients. The study utilizes Federated Averaging (FedAvg~\cite{li2019convergence}), a widely adopted aggregation method for training global models in FL.
\vspace{-0.2cm}
\subsection{Experimental Setup}
\label{sec:experimental-setup}
\noindent \textbf{Applications and Datasets.} Our study analyzes \emph{group-bias} across $K \in \{ 4, 5 \}$ clients (computers that simulate the FL environment, mirroring real-world heterogeneous data collection devices following recent works in FL \cite{hsieh2020non, mohri2019agnostic, yue2021gifair}) using two deep learning models and two datasets. We employ the ResNet model~\cite{he2016deep} for CIFAR10 \cite{krizhevsky2009learning} image classification and a Convolutional Neural Network (CNN) on the DIGITS classification dataset, which comprises data from diverse sources with feature shifts. The goal is to replicate real-world FL scenarios with varied client data. We construct the DIGITS dataset by combining data from SynthDigits 
\cite{ganin2015unsupervised}, MNIST-M \cite{ganin2015unsupervised}, and MNIST \cite{baldominos2019survey}.

We select these datasets to compare \emph{group-bias} with existing bias mitigation techniques in FL. Each dataset is evenly distributed among $K$ clients in the FL framework, ensuring equal allocation of group data points. Clients utilize replicated versions of the original benchmark test set, aligning noise feature distributions between training and test data.

We set all model parameters to match FL parameters for global model convergence under IID data settings, including label and feature noise homogeneity. Client settings include a mini-batch size of $128$, a learning rate of $0.01$, and $40$ (for CIFAR10) and $12$ (for DIGITS) training rounds.

\begin{figure*}
     \centering
      \begin{subfigure}[b]{0.4\linewidth}
         \centering
         \includegraphics[width=\textwidth]{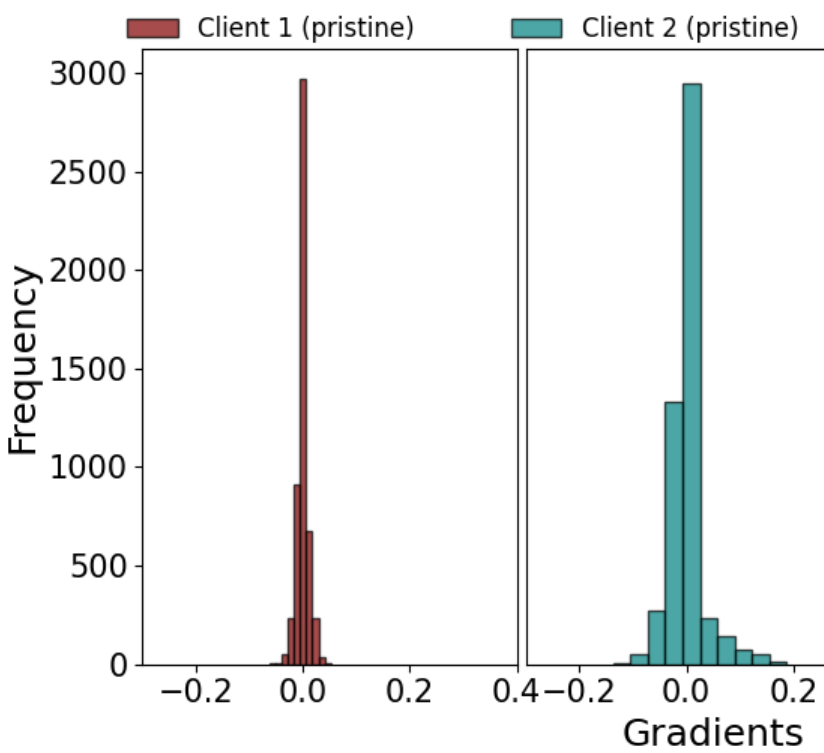}
         \caption{Correlation=0.46}
         \label{fig:grad1}
     \end{subfigure}
     \begin{subfigure}[b]{0.4\linewidth}
         \centering
         \includegraphics[width=\textwidth]{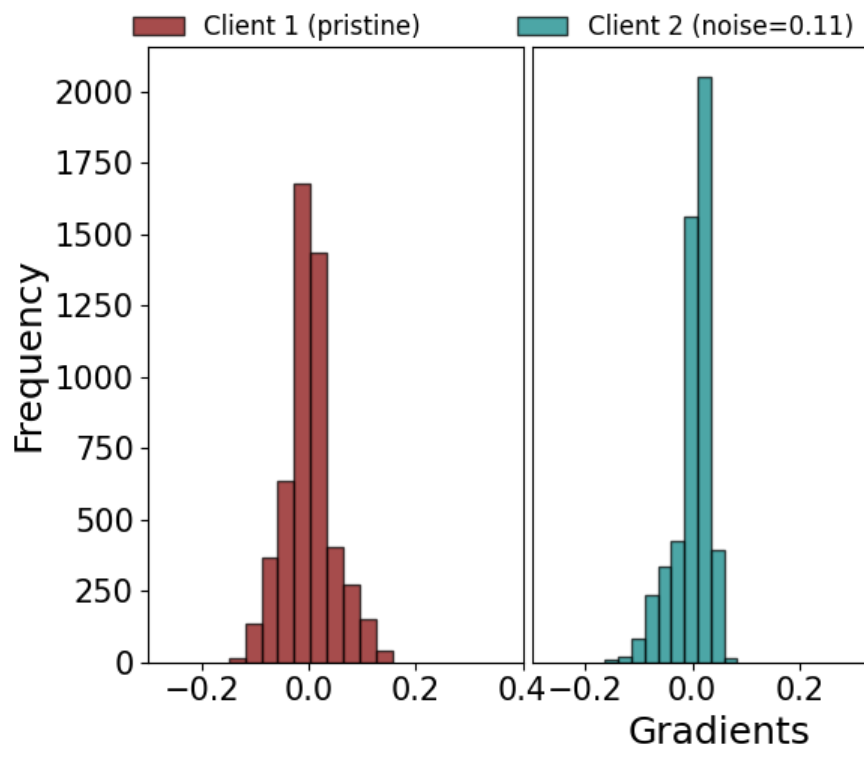}
         \caption{Correlation=-0.14}
         \label{fig:grad2}
     \end{subfigure}
        \caption{Gradient distribution in a fully connected layer on the CIFAR10 dataset. The red and blue bars depict the local gradient distribution on client $1$ and client $2$, respectively. In (a), the distribution of local gradients is demonstrated across the two clients in IID settings. In (b), the distribution is shown in non-IID settings, with the introduction of Gaussian noise with variance $x$ ($\text{noise}=x$) on non-IID clients.}
        \vspace{-0.2cm}
        \label{fig:prelim-grad}
\end{figure*}

\noindent \textbf{Heterogeneous Feature Distributions.} We add noise to mimic real-world distorted images that fail to share the same feature distribution with the pristine training images~\cite{ghosh2018robustness, saenko2010adapting,song2022flair}. In particular, we add Gaussian noise with a variance greater than or equal to $0.03$, consistent with the real-world deployments\cite{lyu2020collaborative}. We create two different distortion levels in each dataset across $K$ clients. For the CIFAR10, three advantaged clients (A, B, C) lack distortions, while the other two disadvantaged clients (D, E) host data with Gaussian noise of variance $var \in \{0.03,0.07,0.11,0.3, 0.4, 0.8, 1.0\}$. For the DIGITS dataset, two advantaged clients (C, D) lack distortions, while the other two disadvantaged clients (A, B) host data with Gaussian noise.
\subsection{Key Findings}
\noindent \textbf{Non-IID Study.} 
We study the FL model's unfairness by examining how the biased global model treats local groups differently for each client. We measure the TPR performance gap between the best and worst groups using each client's local test data (with a similar distortion level as the training data). Figure \ref{fig:cifarweights} shows \emph{group-bias} in CIFAR10, while Figure~\ref{fig:digitsweights} illustrates this in DIGITS. The global model's recognition of local groups varies per client, as seen in the discrepancy between their performances. Increasing Gaussian noise on a client amplifies this difference, indicating that \emph{heterogeneous local features across clients contribute to group bias}.

\noindent \textbf{Limitation of Federated Averaging.}
We empirically investigate how heterogeneous local data distributions affect local model gradients. Post-convergence, we extract gradients from the last linear layer of each local model across two clients. Figure \ref{fig:prelim-grad} shows histograms of these gradients, highlighting variations across clients with heterogeneous features (\ref{fig:grad2}) compared to more consistent distributions in clients with homogeneous features (\ref{fig:grad1}). In \ref{fig:grad1}, a Spearman correlation \cite{myers2004spearman} of $0.46$ indicates strong correlation and uniformity among clients with IID features. Conversely, in Figure~\ref{fig:grad2}, clients with non-IID features show a correlation of $-0.14$, suggesting dissimilarity.

Our non-IID study underscores the challenges in conventional FedAvg schemes, revealing consistently unfair model behavior across distinct applications and datasets. This problem emphasizes the need for bias mitigation methods to alleviate adverse outcomes, including performance degradation in critical applications like medical contexts and the inability to adapt to dynamic heterogeneous environments.

%% file: 4-methodology.tex
The primary objective of our work is to address group bias resulting from feature heterogeneity across clients, all while preventing the leakage of sensitive data. In this section, we formally define our problem and then present our approach to mitigate group bias without substantially degrading the best group performance.
\begin{algorithm}
\begin{algorithmic}[1]
\State  {\textbf{Input:} ${ (\textbf{x}_i, y_i, g_j, c_k) }$, global fairness learning rate $\eta_ \mu$, iteration count $T$, model class $H$.}

\hspace{-3.5em} Let $\epsilon_{g_j} \longleftarrow \frac{1}{|\textbf{G}|} \sum_{(\textbf{x},y) \in g_j}^{} \mathcal{L} (h_{\theta}  (\textbf{x}), y)$; (for each $c_k$)

\State {Initialize $\lambda_{g_j} \longleftarrow  P(\textbf{G} = g_j)$ and $\theta$ randomly.}

\For { {\texttt{$t = 1$ to $T$}}  }
    \For { each client {\texttt{$c_k$ $\in$ $\textbf{C}$}}  }

    \State {\textbf{Compute} $w_{g_j} ^{t} \longleftarrow \frac{\lambda_{g_j}}{P(\textbf{G} = g_j)} $}

    \State {\textbf{Find} $h_{c_k} \longleftarrow \arg \min_{h \in H} \sum_{g}^{}  w_{g_j} ^{t} \cdot \epsilon_{g_j} (h_{c_k}) $}; (for $h_{c_k} \in H$)

    \State {\textbf{Update} $\lambda_{g_j} \longleftarrow \lambda_{g_j} \cdot  \exp(-{{\eta_ \mu} } \cdot \epsilon_{g_j} (h_{c_k})) $}; (Multiplicative Weight Update)

    \State {\textbf{Send} $h_{c_k}$ ($\boldsymbol{\theta}_{c_k}$) to the server.}

    \EndFor
\EndFor

\State \textcolor{black}{Server \textbf{computes:} $\theta \longleftarrow \sum_{c_k \in C}^{} \frac{n_{c_k}}{n} \boldsymbol{\theta}^c_t $}; (FedAvg: $n_{c_k}$--number of data points at client $c_k$; $n$--total data points in FL)

\hspace{-2.5em} \textbf{Output:} Uniform distribution over the set of models
$h_1,...,h_T$ with parameters $\boldsymbol{\theta}_1,...,\boldsymbol{\theta}_T$, respectively
\caption{MW group-fairness in Federated Learning}
\label{alg:mw}
\end{algorithmic}
\end{algorithm}
\setlength{\textfloatsep}{4pt}
\setlength{\intextsep}{0pt}
\subsection{Problem Statement and Workflow}
Our configuration assumes a 4-tuple:
\begin{align}
(\textbf{x}_{1 \leq i \leq |\textbf{X}|}, &\, y_{1 \leq i \leq |\textbf{Y}|}, g_{1 \leq j \leq |\textbf{G}|} c_{1 \leq k \leq |\textbf{C}|}) \nonumber
\end{align}
drawn from distribution $P(\textbf{X}, \textbf{Y}, \textbf{G}, \textbf{C})$. Here, $\textbf{x}_i \in \textbf{X}$ represents training images from a total of $|\textbf{X}|$ images, $y_i \in \textbf{Y}$ corresponds to $|\textbf{Y}|$ targets, $g_j \in \textbf{G}$ denotes group membership (from $|\textbf{G}|$ groups) of $\textbf{x}_i$, and $c_k$ is the client on which $(\textbf{x}_i, y_i)$ resides out of $|\textbf{C}|$ clients. Our primary goal is to derive a global model $h_{\boldsymbol{\theta}}$ (with parameters $\boldsymbol{\theta}$) that mitigates \emph{group bias} for each client, with following objective:
\vspace{-0.2cm}
\begin{equation}
\mathcal{L} (h_{\boldsymbol{\theta}}) = \arg \min_{h} \frac{1}{|\textbf{C}|} \sum_{c_i=1}^{|\textbf{C}|} \mathcal{L}_{c_k} (h_{\boldsymbol{\theta}}(\textbf{x}_{i,k}), y_{i,k})
\label{eq:loss-equation}
\end{equation}
\begin{figure*}[t]
\centering
\includegraphics[width=.9\linewidth]{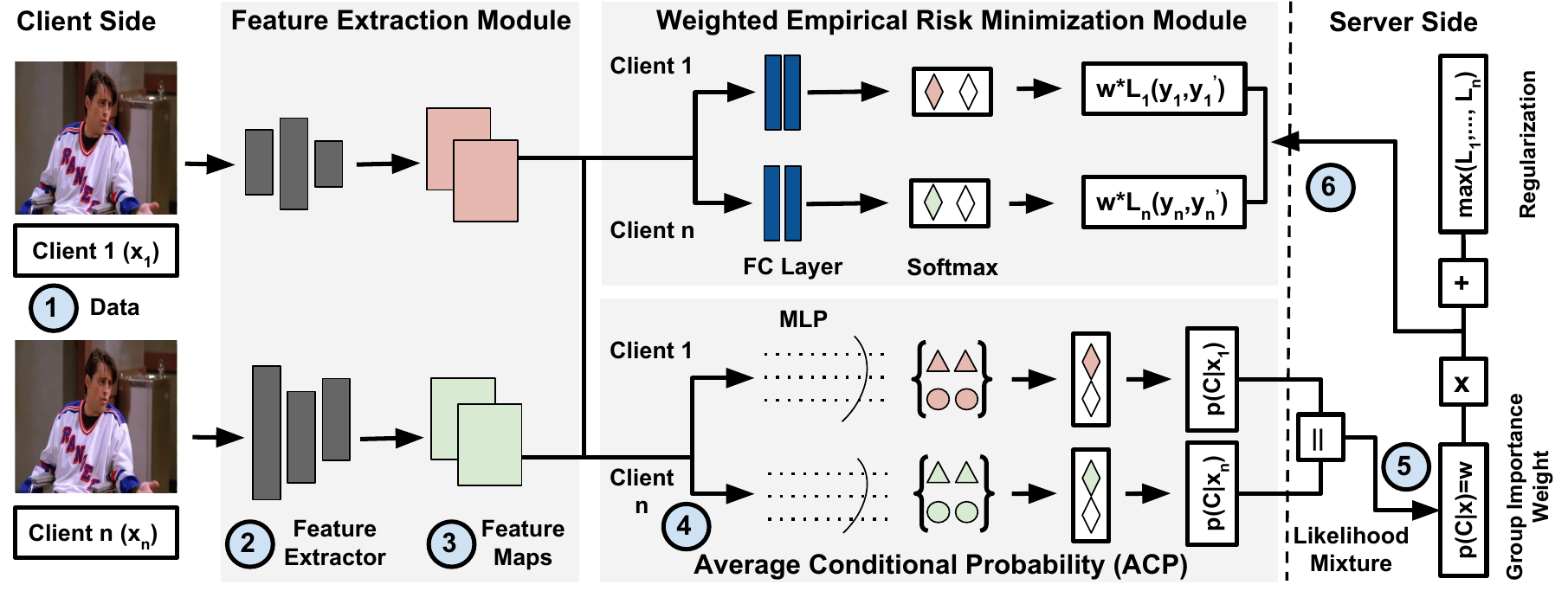}
\vspace{-0.3cm} 
\caption{\textbf{\emph{Overview of the proposed approach.}}}
\label{fig:framework_overview}
\end{figure*}
In equation~\ref{eq:loss-equation} $\mathcal{L}{c_k} (h_{\boldsymbol{\theta}}(\textbf{x}_{i,k}), y_{i,k})$, the empirical risk of client $c_k$ combines group empirical risks $\ell_{g,j} (h(\textbf{x}_{i,k}), y_{i,k})$ with \emph{group importance} $w_{g_j}$. Importance is based on the ratio $\frac{q(g_j|\textbf{x}_i)}{p(g_j|\textbf{x}_i)}$, where $p(g_j|\textbf{x}_i)$ and $q(g_j|\textbf{x}_i)$ represent training and test distributions in $D = \cup D_{c_k}$ (global dataset as a union of local datasets $D_{c_k}$), respectively.
We compute $w_{g_j}$ as an aggregation of all per-client local group importance weights $w_{g_{j,k}}$, $\forall g_j \in \textbf{G}, c_k \in \textbf{C}$. Each $w_{g_j}$ obtained from a multi-class logistic linear regression probabilistic model~\cite{hastie2009elements} is used to train a local model, $h_{\theta_{c_k}}$, minimizing the empirical risk of the \emph{worst-performing} group. On the server side, $h_{\boldsymbol{\theta}_{c_k}}$ from all clients is received and aggregated into a global model $h_{\boldsymbol{\theta}}$.

\noindent\textbf{Workflow.} We illustrate the end-to-end workflow for training with the proposed approach in Figure~\ref{fig:framework_overview}.

\ding{182} In our setup, the server selects all the available clients in each round to avoid the effect of client sampling bias~\cite{coppola2015iterative, zhou2017convergence, wang2019adaptive}. Then, the server distributes copies of the global model to the clients.

\ding{183}-\ding{185} Each client computes the \emph{mixture of group likelihoods}, denoted as $p(g_j | \textbf{x}_i)$ (specifically, $p(g_{j} | \textbf{x}_{i,k})$). In \S~\ref{sec:enabling-fairness}, we outline the privacy-preserving computation details of this denominator, occurring once at the beginning of FL. After each round, clients communicate the local model and local $p(g_{j} | \textbf{x}_{i,k})$ for all groups (only in the first round) to the server.

\ding{186} After clients submit their local models and local $p(g_{j} | \textbf{x}_{i,k})$, the server uses FedAvg to aggregate the local models and generate an updated global model. Additionally, the server computes a \emph{mixture of group likelihoods} for all groups using local likelihoods (emphasizing that this computation occurs once at the beginning of FL).

\ding{187} Each client performs local training after distributing updated global model copies and a \emph{mixture of likelihoods} for all groups. The training involves using our approach \emph{MWR} to adjust group importance weights based on the \emph{mixture of likelihoods} for all groups (\S~\ref{sec:improve-worst}).

\ding{188} Each client computes the performance threshold for the best group and compares it with the best group performance to evaluate \emph{MWR}'s effectiveness in mitigating group bias without compromising the best group performance (\S~\ref{sec:protect-best}).
\subsection{Enabling Privacy-preserving Group Fairness}
\label{sec:enabling-fairness}
Our approach centers on weighting empirical risks with group importance weights, $w_{g_j}$, as shown in Equation~\ref{eq:loss-equation}. Calculating these weights is straightforward in centralized learning~\cite{fang2021learning}, where a global data view is available. However, In FL, lacking this global view is not trivial. We must estimate $w_{g_j}$ while safeguarding client data privacy. Our solution addresses this by approximating the denominator of $w_{g_j}$ ($p(\textbf{G} = g_j | \textbf{X})$) through a process involving a mixture of group likelihoods across clients. Suppose $\textbf{G} = {1,...,j}$ represents groups across clients in FL. Each client $c_k$ employs a multiclass logistic linear regression probabilistic model \cite{abramovich2021multiclass} to predict the likelihood of an input sample $\textbf{x}_{i,k}$ belonging to a specific group $g_j$. The model is defined as $p(\textbf{G} = g_{j}| \textbf{X} = \textbf{x}_{i,k}) = \prod_{j}^{J} f_{\boldsymbol{\theta}, j} (\textbf{x}_i)^{[g_j = j]}$, where $f_{\boldsymbol{\theta}, j} (\textbf{x}_{i,k})^{[g_j = j]}$ is a multinomial probability mass function \cite{kulturel2006multi}. Each client uses the softmax function $f_{\boldsymbol{\theta}, j} (\textbf{x}_{i,k})^{[g_j = j]} = \prod_{j}^{J} \frac{\exp{(\textbf{x}_{i,k} \boldsymbol{\theta}_{c_k})}}{\sum_{j}^{J} \exp{(\textbf{x}_{i,k} \boldsymbol{\theta}_{c_k}})}$ to obtain group membership probabilities ensuring that these probabilities are positive and sum up to one. Clients share their group likelihood estimates with the server. The server then computes each group's global average likelihood using per-client group average likelihood estimates and the \textit{law of total probability}. For an event space $\{ c_1, c_2, ...,c_{|\textbf{C}|}  \}$ with $P(c_k) \geq 0 \quad \forall k$,
\begin{equation}
p(\textbf{G} = g_j | \textbf{X}) = \sum_{j = 1}^{|\textbf{C}|} p(\textbf{G}=g_j | c_k, \textbf{x}_i) p(c_k).
\label{eq:conditional}
\end{equation}
Here $p(\textbf{G}=g_j | c_k, \textbf{x}_i)$ represents per-group likelihood estimates per client, and $p(c_k)$ is the likelihood of a client $c_k$. In our scenario, $p(c_k)$ is uniform for all clients participating in each training round. Utilizing the law of total probability due to independence in clients' participation in FL, the server distributes group likelihood mixtures $p(\textbf{G} = g_j | \textbf{X})$ to all clients. Clients use this information to compute group importance weights $w_{g_j}$, updated using \emph{MWR} in each round based on $p(\textbf{G} = g_j | \textbf{X})$. To ensure data privacy, clients and the server share required information ($p(\textbf{G}=g_j | c_k, \textbf{x}_i)$) by revealing differentially private likelihood estimates.

To solve the group bias problem, we modify the \emph{MW} algorithm and transform it into a constrained optimization problem to improve the performance of the
the worst-performing group.
Algorithm~\ref{alg:mw} details the workings of the \emph{MW} algorithm. We assign each client with groups and a set of $|\textbf{G}|$ classes for the underlying application during the local learning process.
The optimization constraints comprise decisions made by both the local and global models for groups assigned to clients, ensuring fairness in group classification.
Using image features in the training dataset, we validate constraint satisfaction in each local training iteration and identify suitable groups. 
We then associate decisions made by each local model with a group empirical risk that quantifies how well a decision made by the local model satisfies the constraints. Over time, we minimize the overall risk of the global model by ensuring that each local model incurs a low per-group risk.
This involves tracking the global weight for each group and randomly selecting groups with a probability proportional to their importance weights $w_{g_j}$.
In each iteration, we update $w_{g_j}$ using the \emph{MW} algorithm, multiplying their numerator $q(\textbf{G} = g_j | \textbf{G})$ with factors dependent on the risk of the associated group decision. This update is performed while maintaining the denominator $p(\textbf{G} = g_j | \textbf{G})$ fixed as in $\frac{\lambda \cdot \exp{(\eta \cdot  \ell_{g_j} (h)})}{q(\textbf{G} = g_j | \textbf{X})}$, which penalizes costly group decisions. 
\subsection{Ensuring Optimality through Regularization}
\label{sec:improve-worst}
The \emph{MW} algorithm maximizes worst-group performance by scaling the empirical risk and deep neural network weights. However, the weight magnitude does not ensure optimal risk function convergence~\cite{byrd2019effect}. In our setup, model parameters $\theta$ are trained with cross-entropy loss and stochastic gradient descent (SGD) \cite{bottou2010large} optimization,
converging toward the solution of the hard-margin support vector machine\footnote{A linear classification algorithm that seeks a hyperplane with a strict margin, allowing no misclassification in the training data. \cite{sifaou2019phase} } in the direction $\frac{\boldsymbol{\theta}_t}{|| \boldsymbol{\theta}_t ||}$~\cite{soudry2018implicit}. Introducing weight to the loss function may introduce inconsistencies in the margin. Instead of directly applying importance weighting to the empirical risk, we aim to minimize the following objective for each client $k$: $\sum_{c_k = 1}^{|\textbf{C}|} \mathcal{L}_{c_k} (h_{\boldsymbol{\theta}}(\textbf{x}_{i,k}), y_{i,k}) + \frac{\lambda}{m} \sum_{j = 1}^{m} \lVert \boldsymbol{\theta}_{j, c_k } \rVert.$

Since the optimization problem with \emph{importance weighting} is vulnerable to scaling weights and biases, we introduce regularization to the norm of $\theta_{c_k}$ to increase the margin and 
mitigate the risk of its enlargement due to scaling, forming the basis of our \textbf{\emph{Multiplicative Weight update with Regularization (MWR)}} algorithm.
\subsection{Bias Mitigation without Degrading High-Performing Groups}
While \emph{MWR} ensures group fairness,
\emph{importance weighting} approaches may exbibit unfairness by disproportionately focusing on the worst-performing groups, potentially degrading the performance of the best-performing groups in an attempt to reduce the variance in estimating their contributions to the overall performance \cite{diana2021minimax}.
Practically, 
an algorithm for bias mitigation should achieve fairness without significantly degrading the performance of best-performing groups. 
To address this,
we propose a heuristic approach to reweighing the likelihood (\emph{group importance weights}) associated with each data point belonging to group $\textbf{G} = g_j$ in the dataset. Suppose we have a set of unnormalized importance weights $w_1, w_2,...,w_{n}$ corresponding to $n$ data points in a dataset, where each data point has an associated importance weight, we normalize these weights for each group by computing $\hat{w_1}, \hat{w_2},...,\hat{w_{|\textbf{G}|}}$ using:
\begin{equation}
\hat{w_{g_j}} = \frac{ \sum_{i=1}^{n} w_i \mathbb{I}(\textbf{G} = g_j)  }{\sum_{i=1}^{n} w_i}
\label{eq:normalize-equation}
\end{equation}
The rationale behind Equation\ref{eq:normalize-equation} is to distribute emphasis evenly among different groups, preventing a scenario where a single group dominates the estimation due to an excessively high importance weight. Through weight normalization, we ensure that each group's contribution aligns more closely with its true importance or representation within the dataset.
\subsection{Satisfying Performance Thresholds}
\label{sec:protect-best}
Finally, we establish a performance threshold for the best true positive rate (BTPR) to mitigate group bias without significantly compromising the BTPR
. We denote BTPR for a client $c_i$ as $TPR_{best, c_i}$ and WTPR as $TPR_{worst, c_i}$. We define the threshold for the best TPR as $TPR_{threshold}$. Our fairness enforcement objective aims to minimize the gap between the best and worst-performing groups while maintaining a specified level of TPR performance, as follows:
\begin{equation}
TPR_{best, c_i} - TPR_{threshold}  \leq \eta_\mu \times (TPR_{best, c_i} - TPR_{worst, c_i})
\label{eq:threshold-equation}
\end{equation}
\begin{equation}
TPR_{threshold} \geq TPR_{best, c_i} - \eta_\mu \times (TPR_{best, c_i} - TPR_{worst, c_i})
\label{eq:min-threshold-equation}
\end{equation}
Here $\eta_\mu$ is a parameter governing the trade-off between group fairness and performance. Inequality in \ref{eq:threshold-equation} scales the difference between BTPR and WTPR by $\eta_\mu$ and compares it to the difference between the BTPR and the threshold. For each client, we rearrange the inequality in \ref{eq:threshold-equation} to obtain the minimum BTPR threshold as expressed in equation~\ref{eq:min-threshold-equation}.

%% file: 5-evaluation.tex
This section evaluates our \emph{MWR} group-bias mitigation technique on four image classification datasets (CIFAR10, DIGITS, MNIST, and FashionMNIST). We benchmark our approach against standard bias mitigation techniques in FL.
\subsection{Experiment Testbed}
Our evaluation setup uses the same number of clients, data partitioning scheme, and other learning components (such as learning rate, train/test split, batch size, epochs, rounds) described in \S\ref{sec:experimental-setup} unless stated otherwise.

\noindent \emph{\textbf{Baseline.}} 
We evaluate our approach across four key categories, scrutinizing both bias reduction and overall model performance. The \emph{FL baseline category} (FedAvg) represents a conventional learning scheme in FL. In the \emph{FL bias-reduction category}, we include methods such as AFL\cite{mohri2019agnostic}, TERM\cite{li2020tilted}, and GIFAIR-FL~\cite{yue2021gifair}. These methods employ empirical risk reweighting to mitigate bias and adapt the global model to diverse local data distributions. The \emph{FL heterogeneity category} (FedNTD~\cite{lee2022preservation}) specifically addresses performance loss in FL models arising from data heterogeneity by managing global model memory loss. In the \emph{FL robustness category} (SCAFFOLD~\cite{karimireddy2020scaffold}), the focus is on enhancing the resilience of FL models against outliers and noisy data, thereby mitigating the impact of irregularities in specific device local datasets. To ensure a fair evaluation across all baselines, we meticulously calibrate hyperparameters across datasets, guaranteeing the convergence of the global model.

\noindent \emph{\textbf{Hyperparameter Tuning for MWR.}}
We use the same experimental setup as FedAvg, AFL, FedNTD, TERM, GIFAIR-FL, and SCAFFOLD. However, to apply \emph{MWR} update algorithm per-group loss, we set the value of $\eta_ \mu$ (see Algorithm~\ref{alg:mw}) to different values in the set $\{ 0.01, 0.02, 0.001, 0.009, 0.0001 \}$ based on the level of Gaussian noise in data partitions. Finally, \emph{MWR} uses an $L1$ regularization parameter of $0.00001$ for all datasets. 

\begin{table*}[htbp]
\resizebox{\linewidth}{!}{%
\begin{tabular}{|l|c|c|c|c|c|   c|c|c|c|c|  c|c|c|c|c|c|c|c|c|c|}
\hline
\textbf{Algorithms} & \multicolumn{20}{c|}{\textbf{Datasets}} \\ \hline
 & \multicolumn{6}{c|}{\textbf{CIFAR10}} & \multicolumn{4}{c|}{\textbf{DIGITS}} & \multicolumn{5}{c|}{\textbf{Fashion-MNIST}} & \multicolumn{5}{c|}{\textbf{MNIST}} \\ \hline
\multirow{5}{*}{\textbf{FedAvg \cite{li2019convergence}}} &  \textbf{Client \#}& \textbf{1} & \textbf{2} & \textbf{3} & \textbf{4} & \textbf{5} & \textbf{1} & \textbf{2} & \textbf{3} & \textbf{4} & \textbf{1} & \textbf{2} & \textbf{3} & \textbf{4} & \textbf{5} & \textbf{1} & \textbf{2} & \textbf{3} & \textbf{4} & \textbf{5} \\ \cline{2-21} 
 & TPRD $\downarrow$ & 28&	28&	28&	40&	40&	33&	28&	39&	26&	48&	48&	48&	55&	55&	2&	2&	2&	18&	18 \\ \cline{2-21} 
 & TPRSD $\downarrow$ & 9.13&	9.13&	9.13&	13.29&	13.29&	9.01&	9.91&	13.53&	6.1& 14.19&	14.19&	14.19&	16.1&	16.1&	0.6&	0.6&	0.6&	5.29&	5.29 \\ \cline{2-21} 
 & WTPR $\uparrow$ & 64&	64&	64&	54&	54&	63&	66&	53&	69&	47&	47&	47&	40&	40&	98&	98&	98&	74&	74 \\ \cline{2-21} 
 & BTPR $\uparrow$ & 92&	92&	92&	94&	94&	96&	94&	92&	95&	95&	95&	95&	95&	95&	100&	100&	100&	92&	92 \\ \hline
 
\multirow{5}{*}{\textbf{AFL} \cite{mohri2019agnostic}}  
 & TPRD $\downarrow$ & 29&	29&	29&	36&	36&	36&	33&	37&	25&	48&	48&	48&	55&	55&	2&	2&	2&	18&	18 \\ \cline{2-21} 
 & TPRSD $\downarrow$ & 8.82&	8.82&	8.82&	10.3&	10.3&	9.91&	12.04&	12.74& \textbf{5.18}$\odot$& 14.19&	14.19&	14.19&	16.05&	16.05&	0.6&	0.6&	0.6&	5.03&	5.03 \\ \cline{2-21} 
 & WTPR $\uparrow$ & 62&	62&	62&	56&	56&	59&	60&	55&	70&	47&	47&	47&	40&	40&	98&	98&	98&	75&	75 \\ \cline{2-21} 
 & BTPR $\uparrow$ & 91&	91&	91&	92&	92&	95&	93&	92&	95& 95&	95&	95&	95&	95&	100&	100&	100&	93&	93 \\ \hline

 \multirow{5}{*}{\textbf{FedNTD} \cite{lee2022preservation}}  
 & TPRD $\downarrow$ & 26&	26&	26&	36&	36&	27&	28& \textbf{33} $*$ &	28&	46&	46&	46&	50&	50&	2&	2&	2&	17&	17 \\ \cline{2-21} 
 & TPRSD $\downarrow$ & 8.38&	8.38&	8.38&	11.51&	11.51&	8.02&	7.88&	12.54&	7.71& 13.83&	13.83&	13.83&	14.96&	14.96&	0.6&	0.6&	0.6& \textbf{4.24} $\odot$& \textbf{4.24} $\odot$ \\ \cline{2-21} 
 & WTPR $\uparrow$ & 66&	66&	66&	57&	57&	66&	65&	56&	64&	49&	49&	49&	45&	45&	97&	97&	97&	76&	76 \\ \cline{2-21} 
 & BTPR $\uparrow$ & 92&	92&	92&	93&	93&	93&	93&	89&	92&	95&	95&	95&	95&	95&	99&	99&	99&	93&	93 \\ \hline

 \multirow{5}{*}{\textbf{TERM} \cite{li2020tilted}}  
 & TPRD $\downarrow$ & 26&	26&	26&	34&	34&	34&	33&	36&	24&	48&	48&	48&	55&	55&	2&	2&	2&	19&	19 \\ \cline{2-21} 
 & TPRSD $\downarrow$ & 8.02&	8.02&	8.02&	11.32&	11.32&	9.41&	11.32&	13.03&	5.9& 14.06&	14.06&	14.06&	16.11&	16.11&	0.6&	0.6&	0.6&	5.41&	5.41 \\ \cline{2-21} 
 & WTPR $\uparrow$ & \textbf{69}$\bullet$&	\textbf{69}$\bullet$&	\textbf{69}$\bullet$&	61&	61&	61&	61&	54&	70&	47&	47&	47&	40&	40&	98&	98&	98&	74&	74 \\ \cline{2-21} 
 & BTPR $\uparrow$ & 95&	95&	95&	95&	95&	95&	94&	90&	94&	95&	95&	95&	95&	95&	100&	100&	100&	93&	93 \\ \hline

 \multirow{5}{*}{\textbf{GIFAIR-FL} \cite{yue2021gifair}}  
& TPRD $\downarrow$ & \textbf{24} $*$&	\textbf{24$*$} &	\textbf{24$*$} &	36&	36&	26&	30&	48&	39&	43&	43&	43&	50&	50&	2&	2&	2&	16&	16 \\ \cline{2-21} 
 & TPRSD $\downarrow$ & 8.47&	8.47&	8.47&	11.17&	11.17&	7.82&	8.55&	15.63&	13.16&	12.82&	12.82&	12.82&	14.48&	14.48&	0.6&	0.6&	0.6&	5.47&	5.47 \\ \cline{2-21} 
 & WTPR $\uparrow$ & 68&	68&	68&	56&	56&	68&	64&	44&	52&	53&	53&	53&	46&	46&	98&	98&	98&	76&	76 \\ \cline{2-21} 
 & BTPR $\uparrow$ & 92&	92&	92&	92&	92&	94&	94&	92&	91&	96&	96&	96&	96&	96&	100&	100&	100&	92&	92 \\ \hline
 
 \multirow{5}{*}{\textbf{SCAFFOLD} \cite{karimireddy2020scaffold}} 
 & TPRD $\downarrow$ & 29&	29&	29&	65&	65&	60&	64&	84&	73&	50&	50&	50&	60&	60&	2&	2&	2&	25&	25 \\ \cline{2-21} 
 & TPRSD $\downarrow$ & 10.19&	10.19&	10.19&	20.42&	20.42&	18.02&	20.94&	26.35&	24.64&	14.63&	14.63&	14.63&	17.24&	17.24&	1.36&	1.36&	1.36&	6.57&	6.57 \\ \cline{2-21} 
 & WTPR $\uparrow$ & 63&	63&	63&	32&	32&	37&	28&	12&	20&	46&	46&	46&	35&	35&	97&	97&	97&	70&	70 \\ \cline{2-21} 
 & BTPR $\uparrow$ & 92&	92&	92&	97&	97&	97&	92&	96&	93&	96&	96&	96&	95&	95&	99&	99&	99&	95&	95 \\ \hline
 \multirow{5}{*}{\textbf{\emph{MWR}}} 
 & TPRD $\downarrow$ & 25&	25&	25&	\textbf{30} $*$ & \textbf{30} $*$ &	\textbf{21} $*$& \textbf{19} $*$ &	39&	\textbf{23} $*$ &	\textbf{37} $*$&	\textbf{37} $*$&	\textbf{37} $*$&	\textbf{30} $*$&	\textbf{30} $*$&	\textbf{1} $*$&	\textbf{1} $*$&	\textbf{1} $*$&	\textbf{13} $*$& \textbf{13} $*$ \\ \cline{2-21} 
 & TPRSD $\downarrow$ & \textbf{7.94}$\odot$&	\textbf{7.94}$\odot$&	\textbf{7.94}$\odot$& \textbf{10.05}$\odot$&	\textbf{10.05}$\odot$&	\textbf{5.79}$\odot$& \textbf{5.86}$\odot$& \textbf{11.79}$\odot$&	5.9& \textbf{11.02}$\odot$& \textbf{11.02}$\odot$ & \textbf{11.02}$\odot$ & \textbf{11.17}$\odot$&	\textbf{11.17}$\odot$&	\textbf{0.4}$\odot$& \textbf{0.4$\odot$}&	\textbf{0.4}$\odot$&	4.83&	4.83 \\ \cline{2-21} 
 & WTPR $\uparrow$ & 68&	68&	68&	\textbf{63}$\bullet$& \textbf{63}$\bullet$& \textbf{77}$\bullet$& \textbf{77}$\bullet$&	\textbf{58}$\bullet$& \textbf{73}$\bullet$& \textbf{61}$\bullet$&	\textbf{61}$\bullet$& \textbf{61}$\bullet$& \textbf{66}$\bullet$&	\textbf{66}$\bullet$& \textbf{99}$\bullet$& \textbf{99}$\bullet$& \textbf{99}$\bullet$& \textbf{80}$\bullet$& \textbf{80}$\bullet$ \\ \cline{2-21} 
 
 & BTPR-threshold & 92.5 & 92.5 & 92.5 &	92.4 & 92.4 & 97.8 & 95.8 &	96.6 & 95.7 & 97.6 &	97.6 & 97.6 & 95.7 & 95.7 & 99.9 & 99.9 & 99.9 & 92.9 & 92.9 \\ \cline{2-21} 
 
 & BTPR $\uparrow$ & \textbf{93$\rhd$}&	\textbf{93$\rhd$}&	\textbf{93$\rhd$}&	\textbf{93$\rhd$}&	\textbf{93$\rhd$}&	\textbf{98$\rhd$}&	\textbf{96$\rhd$}&	\textbf{97$\rhd$}&	\textbf{96$\rhd$}& \textbf{98$\rhd$}&	\textbf{98$\rhd$}&	\textbf{98$\rhd$}&	\textbf{96$\rhd$}&	\textbf{96$\rhd$}&	\textbf{100$\rhd$}&	\textbf{100$\rhd$}&	\textbf{100$\rhd$}&	\textbf{93$\rhd$}&	\textbf{93$\rhd$} \\ \hline
\end{tabular}%
}
\caption{Performance evaluation of bias mitigation techniques across various datasets and benchmark models under low-grade noise. 
Symbols used:  $\uparrow$indicates that higher values are more desirable, while $\downarrow$ indicates that lower values are more desirable. For each client across each benchmarks in a particular dataset $*$ signifies the best TPRD; $\odot$ designates the best TPRSD; $\bullet$ represents the best WTPR; and $\rhd$ indicates the best BTPR . (Note: On DIGITS dataset,training involves only $4$ clients, reflecting its composition of merely $4$ heterogeneous datasets.) 
} 
\label{tab:low-grade-noise}
\end{table*}

\subsection{Efficacy and Robustness Analysis}
We now assess the efficacy and robustness of our \emph{MWR} group-bias mitigation technique with the baselines.
 \subsubsection{\textbf{Effect on Group Bias}} 
We assess the efficacy of \emph{MWR}'s group-bias mitigation through: (i) evaluating the \textit{best- and worst-group performance (TPR)}, (ii) analyzing the \textit{TPR group variance per client}, and (iii) examining the \textit{TPR discrepancy per client}. This evaluation is conducted on four datasets, incorporating \emph{low-grade distortion} to simulate prevalent real-world heterogeneity~\cite{hsieh2020non}.

\begin{table*}[htbp]
\resizebox{\linewidth}{!}{%
\begin{tabular}{|l|c|c|c|c|c|   c|c|c|c|c|  c|c|c|c|c|c|c|c|c|c|}
\hline
\textbf{Algorithms} & \multicolumn{20}{c|}{\textbf{Datasets}} \\ \hline
 & \multicolumn{6}{c|}{\textbf{CIFAR10}} & \multicolumn{4}{c|}{\textbf{DIGITS}} & \multicolumn{5}{c|}{\textbf{Fashion-MNIST}} & \multicolumn{5}{c|}{\textbf{MNIST}} \\ \hline
\multirow{5}{*}{\textbf{FedAvg \cite{li2019convergence}}} &  \textbf{Client \#}& \textbf{1} & \textbf{2} & \textbf{3} & \textbf{4} & \textbf{5} & \textbf{1} & \textbf{2} & \textbf{3} & \textbf{4} & \textbf{1} & \textbf{2} & \textbf{3} & \textbf{4} & \textbf{5} & \textbf{1} & \textbf{2} & \textbf{3} & \textbf{4} & \textbf{5} \\ \cline{2-21} 
 & TPRD $\downarrow$ & 31&	31&	31&	46&	46&	46&	39&	42&	29&	48&	48&	48&	59&	59&	2&	2&	2&	29&	29 \\ \cline{2-21} 
 & TPRSD $\downarrow$ & 9.91&	9.91&	9.91&	14.7&	14.7&	13.39&	12.48&	13.85&	6.87& 14.39&	14.39&	14.39&	17.05&	17.05&	0.78&	0.78&	0.78&	8.45&	8.45 \\ \cline{2-21} 
 & WTPR $\uparrow$ & 61&	61&	61&	48&	48&	46&	56&	50&	\textbf{66}$\bullet$& 46&	46&	46&	35&	35&	97&	97&	97&	51&	51 \\ \cline{2-21} 
 & BTPR $\uparrow$ & 92&	92&	92&	94&	94&	92&	95&	92&	95&	94&	94&	94&	94&	94&	99&	99&	99&	80&	80 \\ \hline
 
\multirow{5}{*}{\textbf{AFL} \cite{mohri2019agnostic}}  
 & TPRD $\downarrow$ & 33&	33&	33&	44&	44&	47&	43&	40&	29&	49&	49&	49&	59&	59&	2&	2&	2&	28&	28 \\ \cline{2-21} 
 & TPRSD $\downarrow$ & 9.36&	9.36&	9.36&	14.13&	14.13&	13.6&	14.01& \textbf{13.35} $\odot$ &	6.9& 14.65&	14.65&	14.65&	17.11&	17.11&	0.7&	0.7&	0.7&	7.63&	7.63 \\ \cline{2-21} 
 & WTPR $\uparrow$ & 61&	61&	61&	49&	49&	43&	49&	52&	66&	45&	45&	45&	35&	35&	97&	97&	97&	52&	52 \\ \cline{2-21} 
 & BTPR $\uparrow$ & 94&	94&	94&	93&	93&	90&	92&	92&	95&	94&	94&	94&	94&	94&	99&	99&	99&	80&	80 \\ \hline

 \multirow{5}{*}{\textbf{FedNTD} \cite{lee2022preservation}}  
 & TPRD $\downarrow$ & 26&	26&	26&	56&	56&	35&	29&	\textbf{37}& \textbf{27} &46&	46&	46&	50&	50&	2	&2&	2&	25&	25 \\ \cline{2-21} 
 & TPRSD $\downarrow$ & 8.91&	8.91&	8.91&	16.69&	16.69&	10.65&	9.03&	13.64&	8.76& 13.83&	13.83&	13.83&	15.01&	15.01&	0.74&	0.74&	0.74&	6.77&	6.77 \\ \cline{2-21} 
 & WTPR $\uparrow$ & 65&	65&	65&	40&	40&	54&	59&	51&	64&	49&	49&	49&	45&	45&	97&	97&	97&	56&	56 \\ \cline{2-21} 
 & BTPR $\uparrow$ & 91&	91&	91&	96&	96&	89&	88&	88&	91&	95&	95&	95&	95&	95&	99&	99&	99&	81&	81 \\ \hline

 \multirow{5}{*}{\textbf{TERM} \cite{li2020tilted}}  
 & TPRD $\downarrow$ & \textbf{23} $*$& \textbf{23} $*$& \textbf{23} $*$&	40&	40&	47&	40&	43&	30&	48&	48&	48&	59&	59&	2&	2&	2&	30&	30 \\ \cline{2-21} 
 & TPRSD $\downarrow$ & 7.9&	7.9&	7.9&	13.41&	13.41&	13.72&	13.07&	14.13& \textbf{5.87} $\odot$	& 14.39&	14.39&	14.39&	17.08&	17.08&	0.78&	0.78&	0.78&	8.64&	8.64 \\ \cline{2-21} 
 & WTPR $\uparrow$ & \textbf{69}$\bullet$ & \textbf{69}$\bullet$& \textbf{69}$\bullet$&	53&	53&	44&	55&	49&	65&	46&	46&	46&	35&	35&	97&	97&	97&	51&	51 \\ \cline{2-21} 
 & BTPR $\uparrow$ & 92&	92&	92&	93&	93&	91&	95&	92&	95&	94&	94&	94&	94&	94&	99&	99&	99&	81&	81 \\ \hline

 \multirow{5}{*}{\textbf{GIFAIR-FL} \cite{yue2021gifair}} 
 & TPRD $\downarrow$ & 30&	30&	30&	53&	53&	32&	37&	48&	40&	45&	45&	45&	53&	53&	2&	2&	2&	27&	27 \\ \cline{2-21} 
 & TPRSD $\downarrow$ & \textbf{8.16} $\odot$ &	\textbf{8.16} $\odot$ &	\textbf{8.16} $\odot$&	14.92&	14.92&	10.13&	10.18&	15.69&	13.06&	13.4&	13.4&	13.4&	15.46&	15.46&	0.66&	0.66&	0.66&	7.64&	7.64 \\ \cline{2-21} 
 & WTPR $\uparrow$ & 63&	63&	63&	42&	42&	56&	56&	43&	51&	51&	51&	51&	42&	42&	98&	98&	98&	54&	54 \\ \cline{2-21} 
 & BTPR $\uparrow$ & 93&	93&	93&	95&	95&	88&	93&	91&	91&	96&	96&	96&	95&	95&	100&	100&	100&	81&	81 \\ \hline
 
 \multirow{5}{*}{\textbf{SCAFFOLD} \cite{karimireddy2020scaffold}} 
 & TPRD $\downarrow$ & 38&	38&	38&	94&	94&	47&	60&	84&	74&	51&	51&	51&	63&	63&	5&	5	&5&	54&	54 \\ \cline{2-21} 
 & TPRSD $\downarrow$ & 13.21&	13.21&	13.21&	26.53&	26.53&	14.73&	18.13&	27.46&	23.65&	14.77&	14.77&	14.77&	18.27&	18.27&	1.32&	1.32&	1.32&	14.06&	14.06 \\ \cline{2-21} 
 & WTPR $\uparrow$ & 57&	57&	57&	5&	5&	48&	35&	10&	22&	45&	45&	45&	31&	31&	95&	95&	95&	29&	29 \\ \cline{2-21} 
 & BTPR $\uparrow$ & 95&	95&	95&	99&	99&	95&	95&	94&	96&	96&	96&	96&	94&	94&	100&	100&	100&	83&	83 \\ \hline
 
 \multirow{5}{*}{\textbf{\emph{MWR}}} 
 & TPRD $\downarrow$ & 29&	29&	29& \textbf{33} $*$& \textbf{33} $*$&	\textbf{28} $*$& \textbf{24} $*$&	44&	29&	\textbf{38} $*$&	\textbf{38} $*$&	\textbf{38} $*$&	\textbf{34} $*$&	\textbf{34} $*$&	\textbf{2} $*$&	\textbf{2} $*$&	\textbf{2} $*$&	\textbf{20} $*$& \textbf{20} $*$ \\ \cline{2-21} 
 
 & TPRSD $\downarrow$ & 10.29& 10.29&	10.29& \textbf{12.27} $\odot$ &	\textbf{12.27}$\odot$&	\textbf{8.01}$\odot$&	\textbf{8.09}$\odot$&	13.76&	7.98& \textbf{11.35} $\odot$&	\textbf{11.35}$\odot$&	\textbf{11.35}$\odot$&	\textbf{12.44}$\odot$&	\textbf{12.44}$\odot$& \textbf{0.63}$\odot$&	\textbf{0.63}$\odot$&	\textbf{0.63}$\odot$& \textbf{6.45}$\odot$& \textbf{6.45}$\odot$ \\ \cline{2-21} 
 
 & WTPR $\uparrow$ & 66&	66&	66&	\textbf{58}$\bullet$& \textbf{58}$\bullet$& \textbf{68}$\bullet$& \textbf{69}$\bullet$&	\textbf{51}$\bullet$&	65&	 \textbf{59}$\bullet$& \textbf{59}$\bullet$& \textbf{59}$\bullet$& \textbf{62}$\bullet$&	\textbf{62}& \textbf{98}$\bullet$ &	\textbf{98}$\bullet$& \textbf{98}$\bullet$& \textbf{60}$\bullet$& \textbf{90}$\bullet$ \\ \cline{2-21} 
 
 & BTPR-threshold & 94.7 & 94.7 & 94.7 &	90.6 & 90.6 & 95.7 & 92.7 &	94.6 & 93.7 &  96.6 &	96.6 & 96.6 & 95.6 & 95.6 & 99.9 & 99.9 & 99.9 & 79.8 & 78.8 \\ \cline{2-21}
 
 & BTPR $\uparrow$ & \textbf{95$\rhd$}&	\textbf{95$\rhd$}&	\textbf{95$\rhd$}&	\textbf{91$\rhd$}&	\textbf{91$\rhd$}&	\textbf{96$\rhd$}&	\textbf{93$\rhd$}&	\textbf{95$\rhd$}&	\textbf{94$\rhd$}&	\textbf{97$\rhd$}&	\textbf{97$\rhd$}&
 \textbf{97$\rhd$}&	\textbf{96$\rhd$}&	\textbf{96$\rhd$}&	\textbf{100$\rhd$}&	\textbf{100$\rhd$}&	\textbf{100$\rhd$}&	\textbf{80 $\rhd$}&	\textbf{80$\rhd$} \\ \hline
\end{tabular}%
}
\caption{Performance evaluation of bias mitigation techniques across various datasets and benchmark models under low-grade noise. 
Symbols used:  $\uparrow$indicates that higher values are more desirable, while $\downarrow$ indicates that lower values are more desirable. For each client across each benchmarks in a particular dataset $*$ signifies the best TPRD; $\odot$ designates the best TPRSD; $\bullet$ represents the best WTPR; and $\rhd$ indicates the best BTPR . (Note: On DIGITS dataset,training involves only $4$ clients, reflecting its composition of merely $4$ heterogeneous datasets.)} 
\label{tab:high-grade-noise}
\end{table*}
Table\ref{tab:low-grade-noise} presents the TPR, TPRSD, WTPR, and BTPR perfromance scores across various bias mitigation techniques and datasets. Notably, among these techniques, \emph{MWR} stands out by achieving a significantly fairer outcomes for groups. We can see that our algorithm substantially decreases TPRSD across most clients while maintaining a consistently high TPR. Importance weighting, especially when derived from features characteristics, is powerful in mitigating biases caused by feature noise. If the bias is primarily driven by certain features, assigning appropriate weights to these features can help the model focus on relevant information and reduce the impact of noisy features, resulting in more consistent and equitable predictions.

Although AFL and FedNTD occasionally outperform \emph{MWR} in some instances concerning the TPRSD metric as can be seen in DIGITS dataset's client$4$ and MNIST dataset's clients$4$ and $5$, the differences between the results are marginal.
Importance weighting is sensitive to distribution shifts in the feature space. If there are instances where the distribution shifts significantly, the importance weights may not be as effective. On the other hand, techniques such as FedNTD, through knowledge distillation, seem to be more robust to feature noise as it involves transferring knowledge from a more complex model (teacher) to a simpler one (student), potentially leading to better generalization and lower standard deviation in true positive rates across groups. Additionally, it becomes evident from Table \ref{tab:low-grade-noise} that \emph{MWR} results in an increased WTPR for the group with the smallest TPR, accompanied by the smallest TPRD among the evaluated bias mitigation techniques. 

Importance weights derived from image features captures the distinctive characteristics of different groups more effectively than other methods. This adaptability is crucial in mitigating bias since it tailors the mitigation strategy to the specific features and challenges present in each group. Despite TERM appearing to outperform our proposed method for the minimax group fairness metric (WTPR) in CIFAR10 dataset's clients $1$, $2$, and $3$, this can be understood as a consequence of the reduction in TPR among privileged clients lacking local data with distortions. This reduction elevates the lower TPR among disadvantaged clients affected by distortions
Importantly, the differences between the results are marginal, indicating a closely competitive performance between the methods despite this disparity while elevating the \emph{group-fairness} among clients.

\begin{figure*}
     \centering
     \begin{subfigure}[b]{0.45\linewidth}
         \centering
         \includegraphics[width=\textwidth]{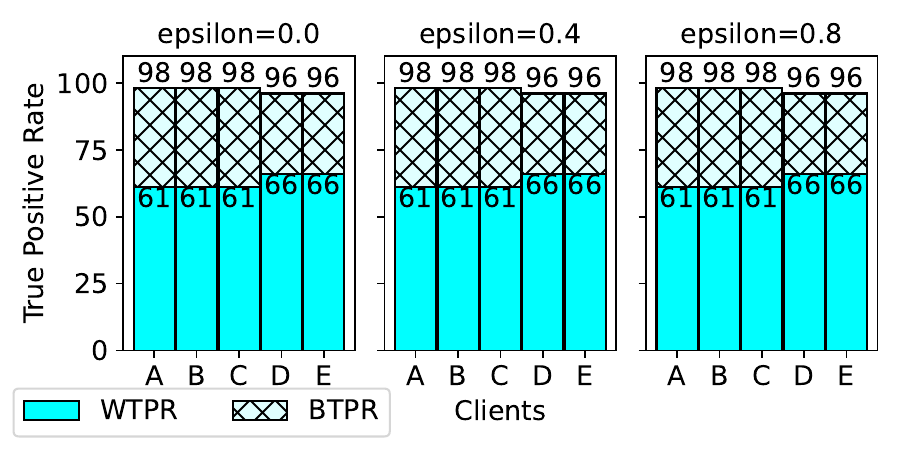}
         \caption{FasionMNIST with DP in MWR ($var=0.3$)}
         \label{fig:privacy-fmnista}
     \end{subfigure}
     \begin{subfigure}[b]{0.45\linewidth}
         \centering
         \includegraphics[width=\textwidth]{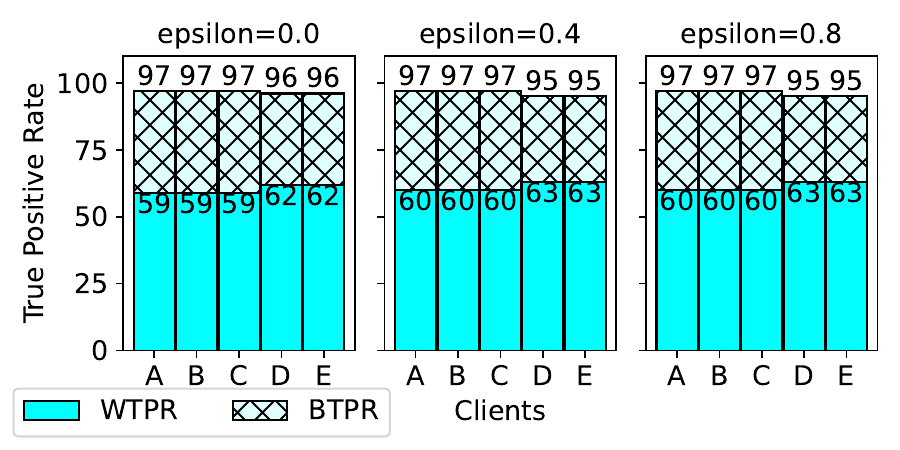}
         \caption{FasionMNIST with DP in MWR ($var=0.4$)}
         \label{fig:privacy-fmnistb}
     \end{subfigure}
        \caption{Examining the performance trade-off in $MWR$ concerning privacy and accuracy across various levels of differential privacy (DP) noise factors on FashionMNIST. In (a), a base Gaussian noise with a variance of $0.3$ is introduced to all methods, while in (b), Gaussian noise with a variance of $0.4$ is applied to all methods. }
        \label{fig:privacy-fmnist}
\end{figure*}

\begin{figure*}[t]
     \centering
     \begin{subfigure}[b]{0.45\linewidth}
         \centering
         \includegraphics[width=\textwidth]{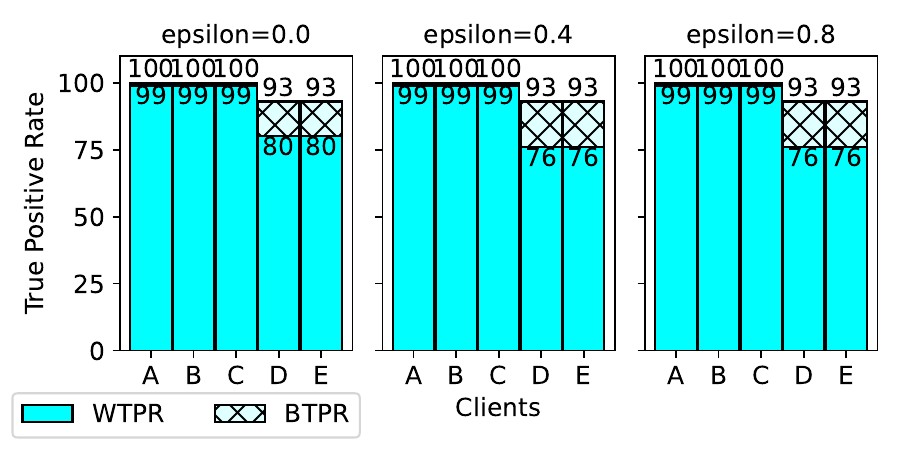}
         \caption{MNIST with DP in MWR ($var=0.8$)}
         \label{fig:fmnist-mwr}
     \end{subfigure}
     \begin{subfigure}[b]{0.45\linewidth}
         \centering
         \includegraphics[width=\textwidth]{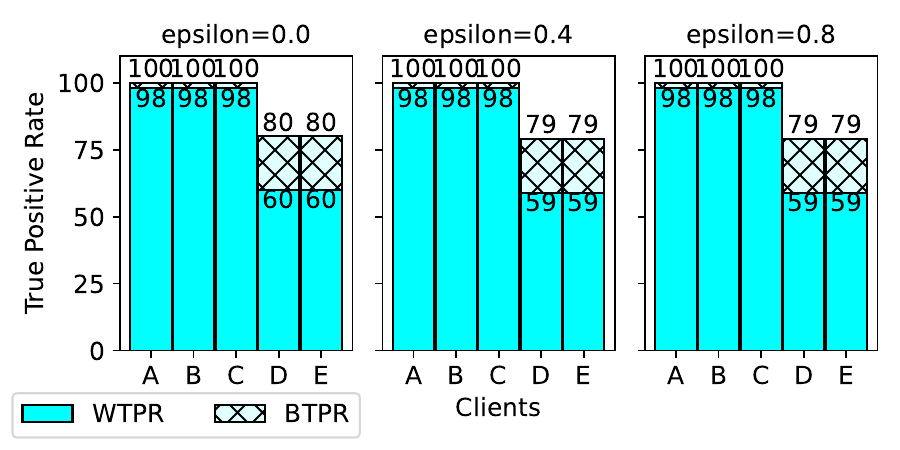}
         \caption{MNIST with DP in MWR ($var=1.1$)}
         \label{fig:fmnist-privacy}
     \end{subfigure}
     \vspace{-0.3cm}
        \caption{Examining the performance trade-off in $MWR$ concerning privacy and accuracy across various levels of differential privacy (DP) noise factors on MNIST. In (a), a base Gaussian noise with a variance of $0.8$ is introduced to all methods, while in (b), Gaussian noise with a variance of $1.1$ is applied to all methods.}
        \label{fig:privacy-mnist}
        \vspace{-0.1cm}
\end{figure*}

\noindent \emph{\textbf{Takeaway:} MWR ensures fairness across groups and maintains predictive accuracy by using importance weights that prioritize the worst-performing groups. Its key strength lies in maintaining fairness without sacrificing performance, achieved through even distribution of importance weights among different groups.}

\subsubsection{\textbf{Robustness of Bias Mitigation}}
In our previous analysis, we added low-grade Gaussian noise to mimic noise in edge device images~\cite{liu2020collabar}. To further test \emph{MWR}'s resilience against increased feature heterogeneity, we raised noise levels in segmented datasets like CIFAR10, MNIST, DIGITS, and Fashion-MNIST to variances of $0.11$, $1.10$, $1.00$, and $0.4$, respectively. Model performance evaluation used the same fairness metrics as before. Table \ref{tab:high-grade-noise} displays TPR, TPRSD, WTPR, and BTPR scores across various bias mitigation techniques and datasets, exploring high-grade distortion scenarios in local data. Consistent with our earlier findings, \emph{MWR} delivers significantly fairer outcomes across diverse groups. The table shows \emph{MWR} reduces TPRSD across most devices while maintaining high TPR. Compared with Table \ref{tab:low-grade-noise}, \emph{MWR} increases WTPR for the lowest TPR group, resulting in minimal TPRD among bias mitigation techniques. This enhancement in WTPR for disadvantaged groups minimally affects high-performing groups' performance.

Although some bias mitigation techniques may slightly outperform in TPRSD and WTPR fairness metrics, this often occurs at the expense of decreased TPR in privileged clients not affected by distortions. However, this decrease compensates for an increase in lower TPR among disadvantaged clients. Despite these differences, the results remain closely competitive among methods, indicating similar performance despite disparity, while simultaneously improving group fairness among clients.

\vspace{0.1cm}
\noindent \emph{\textbf{Takeaway.} our robustness analysis suggests that \emph{MWR} stands out as a robust and fair approach even in scenarios with high-grade heterogeneity, showcasing its effectiveness in mitigating bias across diverse datasets and client groups.}
\subsection{\textbf{Privacy Analysis}}
\label{sec:privacy}
This section explores how differential privacy affects group fairness and performance in \emph{MWR}, particularly in scenarios where local group probability distributions $p(\textbf{G} = g_i | \textbf{x}_{i,k})$ are shared with the server to compute \emph{importance weights}. Differential privacy is crucial for preserving privacy in client metadata, preventing disclosure of sensitive details like group selection probabilities.

We use the MNIST and FashionMNIST datasets for our privacy budget analysis, maintaining consistency in experimental setups and various learning components as detailed in \S\ref{sec:experimental-setup}. We introduce different levels of Laplace noise, denoted by $\epsilon$, to local probability distributions. An $\epsilon$ value of $0.00$ represents perfect differential privacy in the implementation of \emph{MWR}.

\begin{figure*}[t]
     \centering
     \begin{subfigure}[b]{0.45\linewidth}
         \centering
         \includegraphics[width=\textwidth]{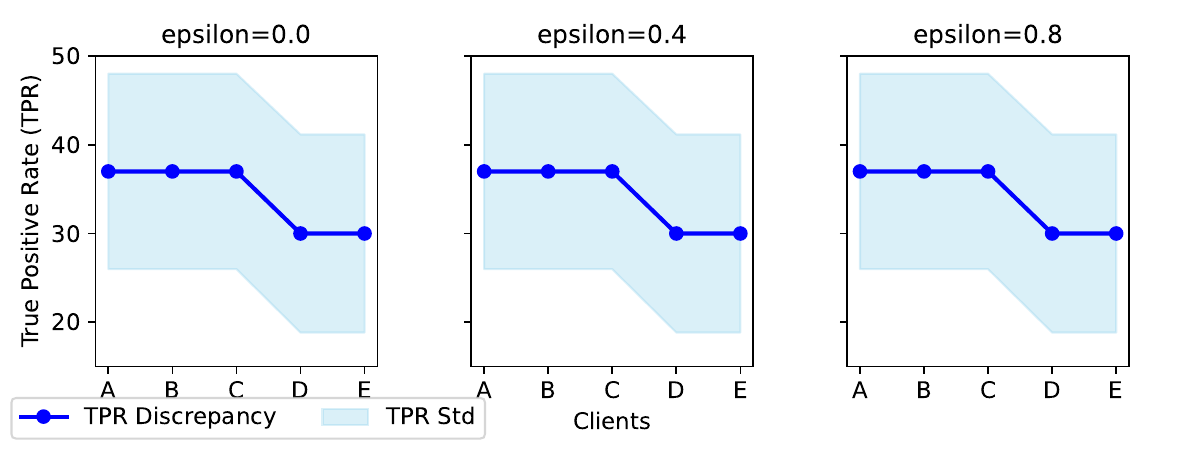}
         \caption{FasionMNIST with DP in MWR ($var=0.3$)}
         \label{fig:privacy-var-fmnista}
     \end{subfigure}
     \begin{subfigure}[b]{0.45\linewidth}
         \centering
         \includegraphics[width=\textwidth]{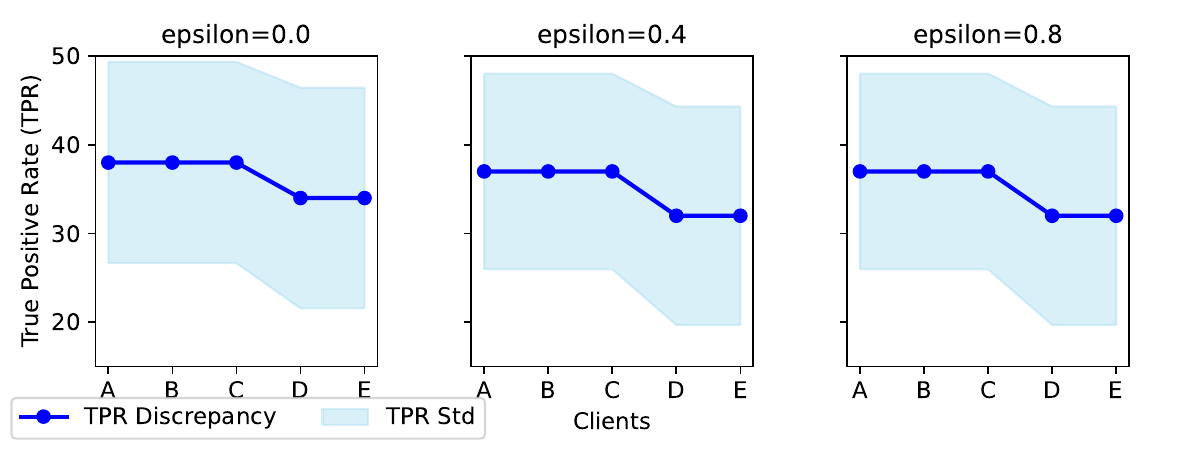}
         \caption{FasionMNIST with DP in MWR ($var=0.3$)}
         \label{fig:privacy-var-fmnistb}
     \end{subfigure}
        \caption{Analyzing the privacy-bias trade-off in $MWR$ across differential privacy (DP) noise levels on FashionMNIST. (a) introduces a base Gaussian noise with a variance of $0.3$, and in (b), Gaussian noise with a variance of $0.4$ is applied. Shaded areas represent deviation represented by TPRSD.}
        \label{fig:privacy-var-fmnist}
        \vspace{-0.1cm}
\end{figure*}

\begin{figure*}
     \centering
     \begin{subfigure}[b]{0.45\linewidth}
         \centering
         \includegraphics[width=\textwidth]{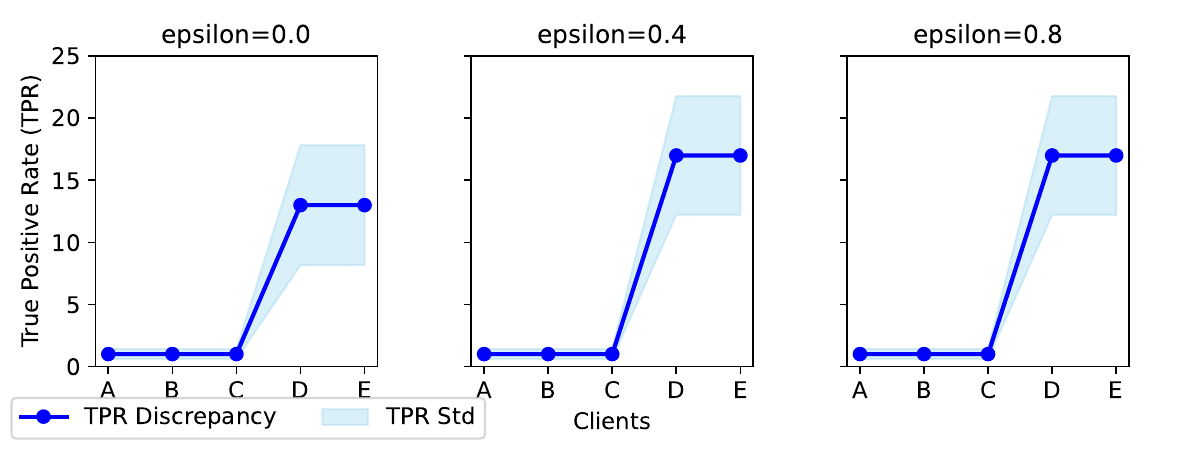}
         \caption{MNIST with DP in MWR ($var=0.8$)}
         \label{fig:privacy-var-mnista}
     \end{subfigure}
     \begin{subfigure}[b]{0.45\linewidth}
         \centering
\includegraphics[width=\textwidth]{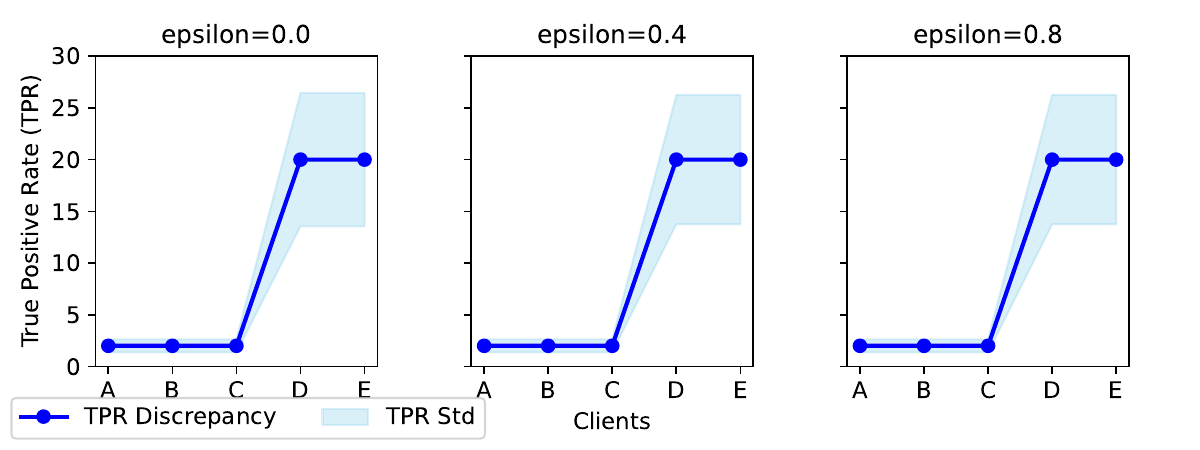}
         \caption{MNIST with DP in MWR ($var=1.1$)}
         \label{fig:privacy-var-mnistb}
     \end{subfigure}
     \vspace{-0.4cm}        \caption{Analyzing the privacy-bias trade-off in $MWR$ across differential privacy (DP) noise levels on MNIST. (a) introduces a base Gaussian noise with a variance of $0.8$, and in (b), Gaussian noise with a variance of $1.1$ is applied. Shaded areas represent deviation represented by TPRSD.}
        \label{fig:privacy-var-mnist}
\end{figure*}

Figures \ref{fig:privacy-fmnist} to \ref{fig:privacy-var-mnist} show the impact of varying levels of Laplace noise ($\epsilon$) on group-fairness metrics (WTPR, TPRSD, and TPRD) and group performance (TPR) in MWR, addressing bias in local data with different levels of feature noise. In Figures \ref{fig:privacy-fmnista} to \ref{fig:privacy-var-fmnistb}, we see that using a privacy budget ($\epsilon \in { 0.0, 0.4, 0.8 }$) for metadata exchange maintains fairness metrics similar to deploying MWR without privacy ($\epsilon \longrightarrow \infty$) on MNIST and FashionMNIST. This is evident from minimal variations in WTPR, TPRSD, and TPRD across all clients (with high and low feature heterogeneity) under all privacy budgets. Moreover, the privacy budget ensures fairness while preserving the best and worst TPR performance. This aligns with the fairness guarantee of MWR, as the privacy budget values ($\epsilon \in { 0.0, 0.4, 0.8 }$) fall within a range that provides algorithmic fairness, as noted in \cite{abay2020mitigating}. Our privacy analysis underscores that our method ensures client privacy through differential privacy on shared metadata without significantly affecting bias or accuracy.

\noindent \emph{\textbf{Takeaway.} MWR demonstrates the feasibility of preserving sensitive information while effectively reducing group bias.}
\subsection{\textbf{Fairness Budget Analysis}}
\label{sec:fairness}
\emph{MWR} incorporates a fairness budget, denoted as $\eta_{\mu}$, to regulate importance weight adjustments for fairness. This control mechanism in \emph{MWR} adjusts importance weights based on past group performance (group loss) for fairness metrics. We assess the impact of $\eta_{\mu}$ on group fairness metrics (WTPR, TPRSD, TPRD) using MNIST and FashionMNIST datasets, setting $\eta_{\mu}$ to different values ($-0.009, -0.003, -0.001, -0.0002$). Tables \ref{subfig:fmnist-budget} and \ref{subfig:mnist-budget} show how the fairness budget $\eta_{\mu}$ affects both group fairness and group performance (TPR) with \emph{MWR}. Increasing $\eta_{\mu}$ values improve fairness guarantees, leading to better WTPR, TPRSD, and TPRD due to faster convergence and adaptation to fairness issues. Conversely, lower $\eta_{\mu}$ values result in more gradual adjustments, slowing down the algorithm's fairness improvements. This experiment is crucial for understanding how adjusting fairness settings impacts outcomes, helping us strike a balance between fairness and the specific fairness parameter we use.

\noindent \emph{\textbf{Takeaway.} Fine-tuning the fairness budget in $MWR$ significantly shapes the degree of fairness. Higher values amplify fairness, while lower values diminish it, underscoring the pivotal role of this parameter in mitigating group bias.}

%% file: 6-conclusion.tex
\begin{table*}[]
\resizebox{\textwidth}{!}{%
\small
\begin{tabular}{|c|cccccccccc|cccccccccc|}
\hline
\multirow{2}{*}{} & \multicolumn{10}{c|}{\textbf{noise variance = 0.3}}                                                                                                                             & \multicolumn{10}{c|}{\textbf{noise variance = 0.4}}                                                                                                                                                                                                                                                                     \\ \cline{2-21} 
                  & \multicolumn{5}{c|}{\textbf{$\eta_{\mu}=-0.003$}}                                                                                                                                                        & \multicolumn{5}{c|}{\textbf{$\eta_{\mu}=-0.009$}}                                                                                      & \multicolumn{5}{c|}{\textbf{$\eta_{\mu}=-0.003$}}                                                                                                                                                        & \multicolumn{5}{c|}{\textbf{$\eta_{\mu}=-0.009$}}                                                                                      \\ \hline
                  \hline
\multicolumn{1}{|l|}{\textbf{Client \#}} & \multicolumn{1}{c|}{\textbf{1}} & \multicolumn{1}{c|}{\textbf{2}} & \multicolumn{1}{c|}{\textbf{3}} & \multicolumn{1}{c|}{\textbf{4}} & \multicolumn{1}{c|}{\textbf{5}} & \multicolumn{1}{c|}{\textbf{1}} & \multicolumn{1}{c|}{\textbf{2}} & \multicolumn{1}{c|}{\textbf{3}} & \multicolumn{1}{c|}{\textbf{4}} & \textbf{5} & \multicolumn{1}{c|}{\textbf{1}} & \multicolumn{1}{c|}{\textbf{2}} & \multicolumn{1}{c|}{\textbf{3}} & \multicolumn{1}{c|}{\textbf{4}} & \multicolumn{1}{c|}{\textbf{5}} & \multicolumn{1}{c|}{\textbf{1}} & \multicolumn{1}{c|}{\textbf{2}} & \multicolumn{1}{c|}{\textbf{3}} & \multicolumn{1}{c|}{\textbf{4}} & \textbf{5} \\ \hline
\textbf{TPRD} $\downarrow$     & \multicolumn{1}{c|}{37}             & \multicolumn{1}{c|}{37}             & \multicolumn{1}{c|}{37}             & \multicolumn{1}{c|}{\textbf{29}}    & \multicolumn{1}{c|}{\textbf{29}}    & \multicolumn{1}{c|}{37}    & \multicolumn{1}{c|}{37}    & \multicolumn{1}{c|}{37}    & \multicolumn{1}{c|}{30}    & 30    & \multicolumn{1}{c|}{38}             & \multicolumn{1}{c|}{38}             & \multicolumn{1}{c|}{38}             & \multicolumn{1}{c|}{\textbf{32}}    & \multicolumn{1}{c|}{\textbf{32}}    & \multicolumn{1}{c|}{38}    & \multicolumn{1}{c|}{38}    & \multicolumn{1}{c|}{38}    & \multicolumn{1}{c|}{34}    & 34    \\ \hline
\textbf{TPRSD} $\downarrow$    & \multicolumn{1}{c|}{\textbf{10.92}} & \multicolumn{1}{c|}{\textbf{10.92}} & \multicolumn{1}{c|}{\textbf{10.92}} & \multicolumn{1}{c|}{\textbf{11.11}} & \multicolumn{1}{c|}{\textbf{11.11}} & \multicolumn{1}{c|}{11.02} & \multicolumn{1}{c|}{11.02} & \multicolumn{1}{c|}{11.02} & \multicolumn{1}{c|}{11.17} & 11.17 & \multicolumn{1}{c|}{\textbf{11.21}} & \multicolumn{1}{c|}{\textbf{11.21}} & \multicolumn{1}{c|}{\textbf{11.21}} & \multicolumn{1}{c|}{\textbf{12.05}} & \multicolumn{1}{c|}{\textbf{12.05}} & \multicolumn{1}{c|}{11.35} & \multicolumn{1}{c|}{11.35} & \multicolumn{1}{c|}{11.35} & \multicolumn{1}{c|}{12.44} & 12.44 \\ \hline
\textbf{WTPR} $\uparrow$     & \multicolumn{1}{c|}{\textbf{61}}    & \multicolumn{1}{c|}{\textbf{61}}    & \multicolumn{1}{c|}{\textbf{61}}    & \multicolumn{1}{c|}{\textbf{68}}    & \multicolumn{1}{c|}{\textbf{68}}    & \multicolumn{1}{c|}{61}    & \multicolumn{1}{c|}{61}    & \multicolumn{1}{c|}{61}    & \multicolumn{1}{c|}{66}    & 66    & \multicolumn{1}{c|}{59}             & \multicolumn{1}{c|}{59}             & \multicolumn{1}{c|}{59}             & \multicolumn{1}{c|}{\textbf{63}}    & \multicolumn{1}{c|}{\textbf{63}}    & \multicolumn{1}{c|}{59}    & \multicolumn{1}{c|}{59}    & \multicolumn{1}{c|}{59}    & \multicolumn{1}{c|}{62}    & 62    \\ \hline
\textbf{BTPR} $\uparrow$     & \multicolumn{1}{c|}{98}             & \multicolumn{1}{c|}{98}             & \multicolumn{1}{c|}{98}             & \multicolumn{1}{c|}{97}             & \multicolumn{1}{c|}{97}             & \multicolumn{1}{c|}{98}    & \multicolumn{1}{c|}{98}    & \multicolumn{1}{c|}{98}    & \multicolumn{1}{c|}{96}    & 96    & \multicolumn{1}{c|}{97}             & \multicolumn{1}{c|}{97}             & \multicolumn{1}{c|}{97}             & \multicolumn{1}{c|}{95}             & \multicolumn{1}{c|}{95}             & \multicolumn{1}{c|}{97}    & \multicolumn{1}{c|}{97}    & \multicolumn{1}{c|}{97}    & \multicolumn{1}{c|}{96}    & 96    \\ \hline
\end{tabular}%
}  
  \caption{Impact of the fairness budget $\eta_{\mu}$ on Fashion-MNIST. A base Gaussian noise with a variance of $0.3$, $0.4$ is introduced to MWR in (a) and (b), respectively. $\uparrow$: Higher is best, $\downarrow$: Lower is best.}
\label{subfig:fmnist-budget}
\end{table*}
\begin{table*}[]
\resizebox{\textwidth}{!}{%
\small
\begin{tabular}{|c|cccccccccc|cccccccccc|}
\hline
\multirow{2}{*}{}      & \multicolumn{10}{c|}{\textbf{noise variance = 0.8}}                                                                                   & \multicolumn{10}{c|}{\textbf{noise variance = 1.1}}                                                         \\ \cline{2-21} 
                       & \multicolumn{5}{c|}{\textbf{$\eta_{\mu}=-0.002$}}                                                                                           & \multicolumn{5}{c|}{\textbf{$\eta_{\mu}= -0.001$}}                                                 & \multicolumn{5}{c|}{\textbf{$\eta_{\mu}=-0.002$}}                              & \multicolumn{5}{c|}{\textbf{$\eta_{\mu}=-0.001$}}                                                          \\ \hline
                       \hline
\multicolumn{1}{|l|}{\textbf{Client \#}} & \multicolumn{1}{c|}{\textbf{1}} & \multicolumn{1}{c|}{\textbf{2}} & \multicolumn{1}{c|}{\textbf{3}} & \multicolumn{1}{c|}{\textbf{4}} & \multicolumn{1}{c|}{\textbf{5}} & \multicolumn{1}{c|}{\textbf{1}} & \multicolumn{1}{c|}{\textbf{2}} & \multicolumn{1}{c|}{\textbf{3}} & \multicolumn{1}{c|}{\textbf{4}} & \textbf{5} & \multicolumn{1}{c|}{\textbf{1}} & \multicolumn{1}{c|}{\textbf{2}} & \multicolumn{1}{c|}{\textbf{3}} & \multicolumn{1}{c|}{\textbf{4}} & \multicolumn{1}{c|}{\textbf{5}} & \multicolumn{1}{c|}{\textbf{1}} & \multicolumn{1}{c|}{\textbf{2}} & \multicolumn{1}{c|}{\textbf{3}} & \multicolumn{1}{c|}{\textbf{4}} & \textbf{5} \\ \hline
\textbf{TPRD} $\downarrow$         & \multicolumn{1}{c|}{1}                 & \multicolumn{1}{c|}{1}                 & \multicolumn{1}{c|}{1}                 & \multicolumn{1}{c|}{13}                & \multicolumn{1}{c|}{13}                & \multicolumn{1}{c|}{1}                 & \multicolumn{1}{c|}{1}                 & \multicolumn{1}{c|}{1}                 & \multicolumn{1}{c|}{13}                & 13                & \multicolumn{1}{c|}{2}                 & \multicolumn{1}{c|}{2}                 & \multicolumn{1}{c|}{2}                 & \multicolumn{1}{c|}{20}                & \multicolumn{1}{c|}{20}                & \multicolumn{1}{c|}{2}                 & \multicolumn{1}{c|}{2}                 & \multicolumn{1}{c|}{2}                 & \multicolumn{1}{c|}{20}                & 20                \\ \hline
\textbf{TPRSD}   $\downarrow$      & \multicolumn{1}{c|}{\textbf{0.3}}               & \multicolumn{1}{c|}{\textbf{0.3}}               & \multicolumn{1}{c|}{\textbf{0.3}}               & \multicolumn{1}{c|}{\textbf{4.33}}              & \multicolumn{1}{c|}{\textbf{4.33}}              & \multicolumn{1}{c|}{0.4}               & \multicolumn{1}{c|}{0.4}               & \multicolumn{1}{c|}{0.4}               & \multicolumn{1}{c|}{4.83}              & 4.83              & \multicolumn{1}{c|}{\textbf{0.53}}              & \multicolumn{1}{c|}{\textbf{0.53}}              & \multicolumn{1}{c|}{\textbf{0.53}}              & \multicolumn{1}{c|}{\textbf{6.39}}              & \multicolumn{1}{c|}{\textbf{6.39}}              & \multicolumn{1}{c|}{0.63}              & \multicolumn{1}{c|}{0.63}              & \multicolumn{1}{c|}{0.63}              & \multicolumn{1}{c|}{6.45}              & 6.45              \\ \hline
\textbf{WTPR}   $\uparrow$       & \multicolumn{1}{c|}{99}                & \multicolumn{1}{c|}{99}                & \multicolumn{1}{c|}{99}                & \multicolumn{1}{c|}{80}                & \multicolumn{1}{c|}{80}                & \multicolumn{1}{c|}{99}                & \multicolumn{1}{c|}{99}                & \multicolumn{1}{c|}{99}                & \multicolumn{1}{c|}{80}                & 80                & \multicolumn{1}{c|}{98}                & \multicolumn{1}{c|}{98}                & \multicolumn{1}{c|}{98}                & \multicolumn{1}{c|}{60}                & \multicolumn{1}{c|}{60}                & \multicolumn{1}{c|}{98}                & \multicolumn{1}{c|}{98}                & \multicolumn{1}{c|}{98}                & \multicolumn{1}{c|}{60}                & 60                \\ \hline
\textbf{BTPR} $\uparrow$         & \multicolumn{1}{c|}{100}               & \multicolumn{1}{c|}{100}               & \multicolumn{1}{c|}{100}               & \multicolumn{1}{c|}{93}                & \multicolumn{1}{c|}{93}                & \multicolumn{1}{c|}{100}               & \multicolumn{1}{c|}{100}               & \multicolumn{1}{c|}{100}               & \multicolumn{1}{c|}{93}                & 93                & \multicolumn{1}{c|}{100}               & \multicolumn{1}{c|}{100}               & \multicolumn{1}{c|}{100}               & \multicolumn{1}{c|}{80}                & \multicolumn{1}{c|}{80}                & \multicolumn{1}{c|}{100}               & \multicolumn{1}{c|}{100}               & \multicolumn{1}{c|}{100}               & \multicolumn{1}{c|}{80}                & 80                \\ \hline
\end{tabular}%
}
\caption{Impact of the fairness budget $\eta_{\mu}$ on the TPR, TPRD, and WTPR on MNIST. A base Gaussian noise with a variance of $0.8$, $1.1$ is introduced to MWR in (a) and (b), respectively. $\uparrow$: Higher is best, $\downarrow$: Lower is best.}
\label{subfig:mnist-budget}
\end{table*}
This study explores FL group bias in decentralized, heterogeneous edge deployments, where devices capture data with diverse features often influenced by noise. Our framework, \emph{MWR}, uses \emph{importance weighting} and \emph{average conditional probabilities} based on data features to improve group fairness in FL across varied local datasets. Heterogeneous features in local group data can bias FL models for minority clients, impacting specific groups on those clients. \emph{MWR} addresses this bias by optimizing worst-performing groups without compromising the best-performing ones compared to other FL methods.
While effective, \emph{MWR} relies on group information to mitigate bias across clients, which can lead to persistent loss discrepancies under severe feature heterogeneity. Future work aims to incorporate methods for estimating and denoising data features to reduce noise without compromising data quality.
\emph{MWR} is highly adaptable and can be extended to complex applications beyond image classification. It can optimize diagnostic outcomes in healthcare datasets, handle multimodal and text-based applications like next-character prediction and image captioning, and mitigate bias in emotion prediction applications within FL settings, ensuring equitable outcomes across diverse groups.